%% file: RPG.tex
\documentclass[twoside]{article}

\usepackage[accepted]{aistats2022}

\usepackage[round]{natbib}

\usepackage[utf8]{inputenc} %
\usepackage[T1]{fontenc}    %
\usepackage{hyperref}       %
\usepackage{url}            %
\usepackage{booktabs}       %
\usepackage{amsfonts}       %
\usepackage{nicefrac}       %
\usepackage{microtype}      %
\usepackage{xcolor}         %

\usepackage{amssymb}
\usepackage{amsthm}
\usepackage{amsmath}
\usepackage{graphicx}
\usepackage{subfigure}
\usepackage{algorithm}
\usepackage{algorithmic}
\usepackage{wrapfig}

\input{math_commands.tex}

\makeatletter
\newtheorem*{rep@theorem}{\rep@title}
\newcommand{\newreptheorem}[2]{%
\newenvironment{rep#1}[1]{%
 \def\rep@title{#2 \ref{##1}}%
 \begin{rep@theorem}}%
 {\end{rep@theorem}}}
\makeatother

\newtheorem{theorem}{Theorem}
\newreptheorem{theorem}{Theorem} %

\newtheorem{assumption}{Assumption}

\newtheorem{corollary}{Corollary}
\newreptheorem{corollary}{Corollary} %

\newreptheorem{lemma}{Lemma} %

\newtheorem{remark}{Remark}

\newcommand{\Actions}{\mathcal{A}}
\newcommand{\States}{\mathcal{S}}
\newcommand{\Prob}{p}

\newcommand{\figwidthone}{0.9\textwidth}
\newcommand{\figwidthtwo}{0.48\textwidth}
\newcommand{\figwidththree}{0.355\textwidth}

\newcommand{\LR}{LR estimator}
\newcommand{\RP}{RP estimator}
\newcommand{\RPs}{RP estimators}

\usepackage[utf8]{inputenc}
\usepackage{pgfplots}
\DeclareUnicodeCharacter{2212}{−}
\usepgfplotslibrary{groupplots,dateplot}
\usetikzlibrary{patterns,shapes.arrows}
\pgfplotsset{compat=newest}

\usepackage{sectsty}%
\usepackage{titlecaps}%

\begin{document}

\twocolumn[
\aistatstitle{Model-free Policy Learning with Reward Gradients}
\aistatsauthor{Qingfeng Lan \And Samuele Tosatto \And Homayoon Farrahi \And A. Rupam Mahmood}
\aistatsaddress{University of Alberta \And University of Alberta \And University of Alberta \And University of Alberta \\ CIFAR AI Chair, Amii}
]

\begin{abstract}
Despite the increasing popularity of policy gradient methods, they are yet to be widely utilized in sample-scarce applications, such as robotics. 
The sample efficiency could be improved by making best usage of available information. 
As a key component in reinforcement learning, the reward function is usually devised carefully to guide the agent. 
Hence, the reward function is usually known, allowing access to not only scalar reward signals but also reward gradients. 
To benefit from reward gradients, previous works require the knowledge of environment dynamics, which are hard to obtain. 
In this work, we develop the \textit{Reward Policy Gradient} estimator, a novel approach that integrates reward gradients without learning a model. 
Bypassing the model dynamics allows our estimator to achieve a better bias-variance trade-off, which results in a higher sample efficiency, as shown in the empirical analysis. 
Our method also boosts the performance of Proximal Policy Optimization on different MuJoCo control tasks.
\footnote{The code is available at \url{https://github.com/qlan3/Explorer/tree/RPG}.}
\end{abstract}

\sectionfont{\MakeUppercase}
\section{Introduction}

Policy gradient methods are increasingly popular in the reinforcement learning (RL) community due to their robustness and ability to deal with continuous action spaces. 
They have been employed in a large variety of domains, including computer games~\citep{ vinyals2019grandmaster,berner2019dota,zha2021douzero,badia2020agent57}, simulated robotic tasks~\citep{haarnoja2018soft,lillicrap2015continuous}, and recommendation systems~\citep{zheng2018drn,zhao2018deep,afsar2021reinforcement}.
Despite their popularity, policy gradient methods still suffer from low sample efficiency, which hinders their applicability to complex real-world situations.
One way to mitigate this inefficiency, is to make best use of prior knowledge into the learning system.

In most real-world applications, reward functions are manually designed~\citep{dulac2019challenges}, and therefore known in advance. 
Most of state-of-the-art techniques, however, do not benefit from such prior knowledge. One way to take full advantage of the reward function is to use its gradient w.r.t.\ the policy parameters.
Attempts to use reward gradients are already observed in prior works, such as Deterministic Value-Policy Gradient (DVPG) algorithm~\citep{cai2020deterministic}, Stochastic Value Gradient (SVG) methods~\citep{heess2015learning}, and Dreamer~\citep{hafner2019dream}. 
However, all these methods require a learned transition model, which is well-known to be a harder task than learning value functions; an inaccurate model may even bring negative effect into training~\citep{van2019use,jafferjee2020hallucinating}. 
In this work, we ask this question: \textit{how can we benefit from reward gradients without learning a model?}

To answer this question, we develop a new policy gradient estimator---the \textit{Reward Policy Gradient} (RPG) estimator---that incorporates reward gradients. 
Specifically, in order to apply reward gradients while avoiding the use of a transition model, our approach incorporates two distinct gradient estimators that are popularly used in model-free learning --- the likelihood ratio (LR) estimator and the reparameterization (RP) estimator. 
Broadly speaking, the \LR{} is generally applicable to both discrete and continuous variables, but often suffers from high variance and thus is frequently used in conjunction with variance reduction techniques \citep{williams1992simple,schulman2017proximal}. 
The \LR{} is also a key component in the policy gradient theorem~\citep{sutton2000policy}. 
On the other hand, the \RP{} exhibits lower variance in practice~\citep{greensmith2004variance}, and its recent utilization in policy gradient methods that improved sample efficiency in several benchmark tasks \citep{haarnoja2018soft,lillicrap2015continuous,fujimoto2018addressing}.

Our RPG estimator combines LR and RP gradients to avoid the need for a functional approximation of the model, while taking full advantage of reward gradients.
Based on this new estimator, we propose a new on-policy policy gradient algorithm---the RPG algorithm---that utilizes and takes advantage of both LR and RP gradients. 
We empirically show that by incorporating reward gradients into the gradient estimator, the bias and variance of estimated gradient decrease significantly. 
To analyze the benefit of reward gradients and the properties of our estimator, we test our algorithm on bandit and simple Markov decision processes, where the ground truth gradient is known in closed-form.
Moreover, we compare RPG with a state-of-the-art actor-critic algorithm---Proximal Policy Optimization---on challenging problems, showing that our algorithm outperforms the baseline algorithm.

In the rest of the paper, we first introduce some background knowledge for the LR and RP estimators. We then move to our major theoretical result --- the Reward Policy Gradient theorem. Then we present RPG algorithm as well as the experimental results.

\section{Background}

Consider a Markov decision process (MDP), $M=(\States, \Actions, \Prob, \Prob_0, r, \gamma)$, where $\States$ is the continuous state space, $\Actions$ is the continuous action space, $\Prob: \States \times \Actions \times \States \rightarrow [0, \infty)$ is the state transition probability distribution, $\Prob_0: \States \rightarrow [0, \infty)$ is the initial state probability distribution, $r: \States \times \Actions \rightarrow \real$ is the reward function,\footnote{For simplicity, we assume $r$ is a deterministic function. In the stochastic case, $r$ can be defined as the expected reward for a given state-action pair.} and $\gamma \in [0,1)$ is the discount factor.
In a given MDP $M$, an agent interacts with the environment to generate a trajectory based on a policy distribution $\pi: \States \times \Actions \rightarrow [0,\infty)$. Specifically, the agent starts at some state $S_0\sim \Prob_0(\cdot)$; at each time-step $t = 0, 1, 2, \dots$, the agent samples an action $A_t \in \Actions$ according to the policy $\pi$ (i.e. $A_t \sim \pi(\cdot|S_t)$), receives the immediate reward $R_t = r(S_t, A_t)$, and observes the next state $S_{t+1}$ sampled from the transition probability distribution (i.e. $S_{t+1} \sim \Prob(\cdot | S_t, A_t)$). A trajectory up to time-step $T$ is defined as $\tau_T = (S_0, A_0, R_1, S_1, \cdots, S_T)$. We define the total discounted reward from time-step $t$ over $\tau_T$ as return $G_t = \sum_{k=t}^{T-1} \gamma^{k-t} r(S_k, A_k)$. Value functions are defined to be the expected return under policy $\pi$, $v_{\pi}(s) = \E_{\pi}[G_t | S_t=s]$; similarly, action-value functions are defined as $q_{\pi}(s, a) = \E_{\pi}[G_t | S_t=s, A_t=a]$.

Furthermore, the state-values and action-values can be related as $q_{\pi}(s,a) = r(s,a) + \gamma \int p(s'|s,a) v_{\pi}(s') \mathrm{d}s'$ and $v_{\pi}(s) = \int \pi(a | s) q_{\pi}(s,a) \mathrm{d}a$. The goal of the agent is to obtain a policy $\pi$ that maximizes the expected return starting from initial states. Let a policy $\pi_{\theta}$ be a differentiable function of a weight vector $\theta$. Our goal is to find $\theta$ that maximizes
\begin{align} \label{eq_obj}
J(\theta) = \int p_0(s) v_{\pi_{\theta}}(s) \mathrm{d}s.
\end{align}
To this end, we can apply gradient ascent techniques.
Since the true gradient $\nabla_{\theta} J(\theta)$ is not typically available, we resort to Monte Carlo methods~\citep{mohamed2020monte}. This gradient estimation problem can be formalized as computing the unbiased gradient of an expectation of a function with respect to some parameters of a distribution.
Specifically, let $p_{\theta}(x)$ be the probability distribution of $x$ with parameters $\theta$.~Define $F(\theta) = \int p_{\theta}(x) \phi(x) \mathrm{d}x$. Then the key step of the problem can be formally defined as estimating $\nabla_{\theta} F(\theta)$:
\begin{align}
\nabla_{\theta} F(\theta) = \nabla_{\theta} \E_{X \sim p_{\theta}}[\phi(X)].
\end{align}
The gradient estimation problem is fundamental in many machine learning areas, such as reinforcement learning~\citep{williams1992simple}, variational inference~\citep{hoffman2013stochastic}, evolutionary algorithms~\citep{conti2018improving}, and variational auto-encoders~\citep{kingma2013auto}. In general, there are many approaches for gradient estimation, such as likelihood-ratio gradient estimators, reparameterization gradient estimators, finite difference methods, and mean-valued derivative methods~\citep{pflug1989sampling,carvalho2021empirical}. Next, we introduce two major approaches for solving this problem.

\subsubsection*{Likelihood-Ratio Gradient Estimators}

The likelihood-ratio (LR) gradient estimator~\citep{glynn1990likelihood} is one of the most popular gradient estimators. 
This estimator applies the log-derivative technique $\nabla_{\theta} \log{p_{\theta}(x)} = \frac{\nabla_{\theta} p_{\theta}(x)}{p_{\theta}(x)}$ to obtain the unbiased gradient estimations:
\begin{align}
&\nabla_{\theta} F(\theta) = \nabla_{\theta} \E_{p_{\theta}(x)}[\phi(x)]
= \nabla_{\theta} \int p_{\theta}(x) \phi(x) \mathrm{d}x \nonumber \\
&= \int \phi(x) \nabla_{\theta} p_{\theta}(x) \mathrm{d}x 
= \int \phi(x) p_{\theta}(x) \nabla_{\theta} \log{p_{\theta}(x)} \mathrm{d}x \nonumber \\
&= \E_{X \sim p_{\theta}}[\phi(X) \nabla_{\theta} \log{p_{\theta}(X)}].
\end{align}

The \LR{} is a fundamental component of the policy gradient theorem~\citep{sutton2011reinforcement}.
Partly because of its generality of being applicable to both continuous and discrete action spaces, the \LR{} has been used in many policy gradient methods in RL.
Many actor-critic algorithms also use this estimator to estimate gradient, such as Asynchronous Advantage Actor-Critic~\citep{mnih2016asynchronous}, Trust Region Policy Optimization~\citep{schulman2015trust}, Proximal Policy Optimization (PPO)~\citep{schulman2017proximal}, and Actor-Critic with Experience Replay~\citep{wang2016sample}.

In practice, variance reduction~\citep{greensmith2004variance} is usually necessary to fully exert the power of this estimator. Subtracting a baseline~\citep{williams1992simple} from $\phi(x)$, applying eligibility traces~\citep{singh1996reinforcement}, and utilizing the Generalized Advantage Estimator (GAE)~\citep{schulman2016high} are three effective approaches to mitigate the variance issue of policy gradient based on the \LR{}.

\subsubsection*{Reparameterization Gradient Estimators}

The reparameterization (RP) gradient estimator is also known as the pathwise gradient estimator or the reparameterization technique~\citep{mohamed2020monte}. Given the underlying probability distribution $p_{\theta}(x)$, this estimator takes the advantage of the knowledge of distribution $p_{\theta}(x)$ and reparameterize $p_{\theta}(x)$ with a simpler base distribution $p(\eps)$ that makes two equivalent sampling processes:
\begin{align}
X \sim p_{\theta}(\cdot) \iff X = f_{\theta}(\eps), \, \eps \sim p(\cdot),
\end{align}
where $f_{\theta}$ is a function that maps $\eps$ to $x$. In other words, there are two equivalent ways to sample $X \sim p_{\theta}(x)$: one is to sample it directly; the other way is to first sample $\eps$ from a base distribution $p(\eps)$ and then apply a function $f$ to transform $\eps$ to $X$. For example, assume $X$ is a Gaussian variable where $X \sim \gN(\mu, \sigma)$ and $\theta=[\mu, \sigma]$. Let the base distribution be $p(\eps) = \gN(0,1)$. Then $X$ can be reparameterized as $X = f_{\theta}(\eps) = \mu + \sigma \eps$. For many common continuous distributions (e.g., the Gaussian, Log-Normal, Exponential, and Laplace), there exists such ways to reparameterize them with a simpler base distribution. Finally, we can write the gradient estimation as: 
\begin{align}
&\nabla_{\theta} F(\theta) 
= \nabla_{\theta} \int p_{\theta}(x) \phi(x) \mathrm{d}x \nonumber \\
&= \nabla_{\theta} \int p(\eps) \phi(f_{\theta}(\eps)) \mathrm{d}\eps
= \int p(\eps) \nabla_{\theta} \phi(f_{\theta}(\eps)) \mathrm{d}\eps.
\end{align}

Note that $p_{\theta}(x) = p(\eps) |\nabla_{\eps} f_{\theta}(\eps)|^{-1}$ due to integration by substitution. \RPs{} are only applicable to known and continuous distributions.\footnote{An application to discrete random variables is possible by reparameterizing the Gumble distribution---the continuous counterpart of the categorical distribution \citep{jang2016categorical}.} In general, there is no guarantee that RP outperforms LR~\citep{gal2016uncertainty,parmas2018pipps}. 
However, \RPs{} have a lower variance under certain assumptions~\citep{xu2019variance}, which usually leads to great benefits in many areas~\citep{mohamed2020monte}. 
\citet{kingma2013auto} applied the \RP{} to obtain a differentiable estimator of the variational lower bound that can be optimized directly using standard stochastic gradient methods. \citet{rezende2014stochastic} used \RPs{} for deep generative models and proposed Deep Latent Gaussian Models that are able to generate realistic images. 

In many RL algorithms, \RPs{} play a crucial role to reparameterize actions and decouple the randomness from a policy, such as Soft Actor-Critic (SAC)~\citep{haarnoja2018soft}, SVG~\citep{heess2015learning}, and RELAX~\citep{grathwohl2018backpropagation}. Deterministic Policy Gradient algorithm (DPG)~\citep{silver2014deterministic}, and Deep Deterministic Policy Gradient algorithm (DDPG)~\citep{lillicrap2015continuous}.
Deterministic Value-Policy Gradient algorithm~\citep{cai2020deterministic} can also be viewed as special cases of these algorithms, where the probability density function of the base distribution is a Dirac delta function. Finally, \citet{wang2019generalization} proposed a class of RL algorithms called Reparameterizable RL, where the randomness of the environment is decoupled from the trajectory distribution via the reparameterization technique.

\section{Reward Policy Gradient Theorem}

In this section, we first present our main theoretical result---the \textit{Reward Policy Gradient Theorem}---a new policy gradient theorem that incorporates the gradient of the reward function without using a state transition function explicitly. The reward policy gradient theorem requires perfect knowledge of the reward and the value function. 
We show that an unbiased estimate can be obtained using approximated reward and state-value functions, by defining a set of compatible features, similarly to~\citet{sutton2000policy}.

We begin by assuming that the action $A$ is sampled from the policy $\pi$ parameterized with $\theta$ given the current state $S$: $A \sim \pi_{\theta}(\cdot | S)$. We reparameterize the policy with a function $f$, $A = f_{\theta}(\eps;S), \, \eps \sim p(\cdot)$. Let function $g$ be the inverse function of $f$, that is, $\eps = g_{\theta}(A;S)$ and $A = f_{\theta}(g_{\theta}(A;S);S)$. Furthermore, following~\citet{imani2018off}, we make two common assumptions on the MDP.

\begin{assumption}\label{assum1}
$\States$ and $\Actions$ are closed and bounded.
\end{assumption}

\begin{assumption}\label{assum2}
$p(s' | s,a)$, $f_{\theta}(\eps; s)$, $g_{\theta}(a | s)$, $\pi_{\theta}(a|s)$, $p(\eps)$, $r(s,a)$, $p_0(s)$ and their derivatives are continuous in all variables $s$, $a$, $s'$, $\theta$, and $\eps$.
\end{assumption}

\begin{remark}
The two assumptions above allow us to exchange derivatives and integrals, and the order of multiple integrations, using Fubini's theorem and Leibniz integral rule.
\end{remark}

We use the same objective $J(\theta)$ as in Eq.~\ref{eq_obj}. Then under these two assumptions, we have the following results. The details of all proofs can be found in the appendix.

\begin{theorem}[Reward Policy Gradient]\label{thm_rpg}
Suppose that the MDP satisfies Assumption~\ref{assum1} and \ref{assum2}, then
\begin{align*}
&\nabla_{\theta} J(\theta) \\
&= \int\!\!\!d^{\pi_{\theta}}\!(s) \pi_{\theta}(a|s) p(s'|s,a)\!\! \left[\nabla_{\theta} r(s,f_{\theta}(\eps; s))|_{\epsilon=g_{\theta}(a;s)} \right. \\
&\left. + \gamma v_{\pi_{\theta}}(s') \nabla_{\theta} \log{\pi_{\theta}(a | s)} \right] \mathrm{d}s \mathrm{d}a \mathrm{d}s',
\end{align*}
where $d^{\pi_{\theta}}(s') = \int \sum_{t=0}^{\infty} \gamma^t p_0(s) p(s \to s', t, \pi_{\theta}) \de s$ is the (discounted) stationary state distribution for policy $\pi_{\theta}$ and $p(s \to s', t, \pi_{\theta})$ is the transition probability from $s$ to $s'$ with $t$ steps under policy $\pi_{\theta}$.
\end{theorem}

\textit{Proof sketch of Theorem~\ref{thm_rpg}.} We first apply the policy gradient theorem~\citep{sutton2000policy} and get an intermediate result in form of the action-value function. Next, we split the action-value function in the policy gradient theorem into two parts: the immediate reward and the state-value of the next state. To incorporate reward gradients, we use the RP technique to the immediate reward part; to avoid the knowledge of the model, we apply the LR estimator to the state-value part. Finally, we combine both parts into an unbiased gradient estimation. 

This theorem provides a new way of computing the objective gradients. 
Specifically, it presents the objective gradient in terms of both LR and RP gradients as additive components.
The theorem also presents the first model-free unbiased gradient estimator of the objective function that utilizes gradients of the reward function. Some algorithms also estimate the gradient of the objective using gradients of the reward function, such as DVPG~\citep{cai2020deterministic} and SVG~\citep{heess2015learning}. However, they are model-based algorithms in the sense that they require the knowledge of the state transition function to estimate an unbiased gradient, while our theorem points out a model-free approach without the knowledge of the state transition.

\begin{theorem}[Reward Policy Gradient with Function Approximation] \label{thm_rpg_appro}
Consider a parametric approximation of the reward function $\hat{r}_\omega(s, a)$ and a parametric approximation of the value function $\hat{v}_\phi(s)$ such that
\begin{align*}
&\nabla_a \hat{r}_\omega(s, a) = \nabla_\theta^\intercal f_\theta(\epsilon; s)|_{\epsilon = g_{\theta}(a; s)} \omega, \\
&\int d^{\pi_{\theta}}(s) \pi_{\theta}(a|s) p(s'|s,a) \nabla_{\phi} \hat{v}_{\phi}(s')  \de s \de a \\
=& \int d^{\pi_{\theta}}(s) \pi_{\theta}(a|s) p(s'|s,a) \nabla_{\theta} \log \pi_{\theta}(a|s) \de s \de a,
\end{align*}
where $\phi = \argmin_{\phi}{\E}\left[(\hat{v}_\phi(s') - v(s'))^2\right]$ and
\begin{align*}
\omega = \argmin_{\omega} \sum_{i}\E &\left[\left(\frac\partial{\partial\theta_i} \hat{r}_\omega(s, a) - \frac\partial{\partial\theta_i}r(s, a)\right)^2\right].  
\end{align*}
Then,
\begin{align*}
 \nabla_\theta J(\theta)
= \E & \left[\nabla_\theta\hat{r}_{\omega}(s, f_\theta(\epsilon; s))|_{\epsilon = g_\theta(a; s)} \right. \\
& \left. + \gamma \hat{v}_{\phi}(s')\nabla_\theta \log \pi_{\theta}(a | s) \right].
\end{align*}
Note that all expectations are computed under $s \sim d^{\pi_{\theta}}$, $a \sim \pi_\theta(\cdot | s)$, $s' \sim p(\cdot|s,a)$.
\end{theorem}
By this theorem, we show that when the reward function and the state-value function are approximated by sufficiently good function approximators (e.g., neural networks), we can obtain an unbiased gradient estimation under certain assumptions. The functions $\hat{r}_\omega$ and $\hat{v}_\phi$ are also known as compatible approximators~\citep{sutton2000policy,peters2008reinforcement}. Note that $\phi$, $\omega$, and $\theta$ have same dimensions.

\section{A Reward Policy Gradient Algorithm based on PPO}

\begin{algorithm}[tb]
\caption{RPG} \label{algo_rpg}
\begin{algorithmic}
    \STATE Input: initial policy parameters $\theta$, value estimate parameters $\phi$, and reward estimate parameters $w$.
    \FOR{$k=1,2,\dots$}
        \STATE Collect trajectories $\mathcal{D}=\{\tau_{i}\}$ with policy $\pi_{\theta}$. \\
        \STATE Compute $G_t$ and $G_{t}^{\lambda} = H_{t}^{\text{GAE}(\lambda)} + \hat{v}_{\phi}(S_t)$. \\
        \STATE Compute PPO advantage $H_{t}=H_{t}^{\text{GAE}(\lambda)}$ and normalize.
        \FOR{epoch $=1,2,\dots$}
        \STATE Slice trajectories $\mathcal{D}$ into mini-batches.
            \FOR{each mini-batch $B$}
                \STATE Set $\hat{\rho}_t(\theta)$ by Eq.~\ref{eq_clip_new} and detach it from the computation graph.
                \STATE Reparameterize the action $A_t=f_{\theta}(\epsilon_t; S_t)$.
                \STATE Compute the predicted reward $\hat{R}_{t+1} = \hat{r}_{w}(S_t, f_{\theta}(\epsilon_t; S_t))$. \\
                \STATE Update $\theta$ by maximizing $\E_{B}[\hat{\rho}_t(\theta) H_{t}^{\text{RPG}}]$,
                where $H_{t}^{\text{RPG}} = \hat{R}_{t+1} + (\gamma G_{t+1}^{\lambda} - \hat{v}_{\phi}(S_t))\times\log{\pi_{\theta}(A_t|S_t)}$.
                \STATE Update $\phi$ by minimizing $\E_{B}[(\hat{v}_{\phi}(S_{t})-G_t)^{2}]$.
                \STATE Update $w$ by minimizing $\E_{B}[(\hat{r}_{w}(S_{t}, A_{t})-R_{t+1})^{2}]$.
            \ENDFOR
        \ENDFOR
    \ENDFOR
\end{algorithmic}
\end{algorithm}

Based on the reward policy gradient theorem, many different policy gradient algorithms can be developed that benefit from reward gradients without using a transition model. In this section, we develop a new policy gradient algorithm based on an existing deep policy gradient method called Proximal Policy Optimization (PPO). First, we introduce the baseline subtraction technique which is generally used in policy gradient algorithms to reduce variance. To be specific, we subtract the baseline $v_{\pi_{\theta}}(s)$ from $\gamma v_{\pi_{\theta}}(s')$ in the RPG estimator:
\begin{align}
\nabla_{\theta} r(s,f_{\theta}(\eps; s)) + (\gamma v_{\pi_{\theta}}(s') - v_{\pi_{\theta}}(s)) \nabla_{\theta} \log{\pi_{\theta}(a | s)}. \label{baseline}
\end{align}
Note that no bias is introduced in this step~\citep{sutton2011reinforcement}.

In practice, when the state-value function and the reward function are unknown to the agent, we could use function approximators (e.g., neural networks) to approximate them. Furthermore, we use $\lambda$-return~\citep{sutton2011reinforcement} $G_{t+1}^{\lambda}$ to replace $v_{\pi_{\theta}}(s')$ in Eq.~\ref{baseline}, which has a close relationship to GAE $H_t^{\text{GAE}(\lambda)}$~\citep{schulman2016high}, i.e. $G_{t}^{\lambda} = H_{t}^{\text{GAE}(\lambda)} + v_{\pi_{\theta}}(s_t)$, where $\lambda \in [0,1]$. 
Using $\lambda$-returns significantly reduce the variance of gradient estimations while retaining tolerable biases~\citep{schulman2016high}.

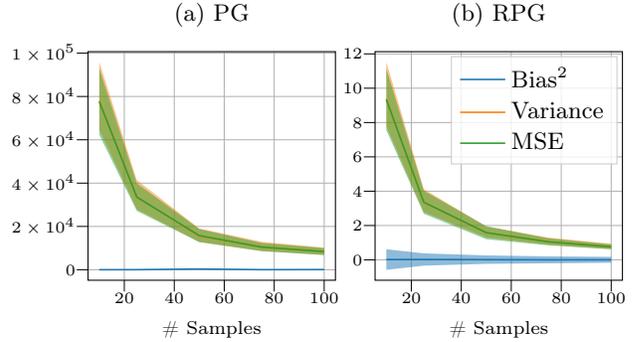
\begin{figure}[t]
\input{figures/lqg.tex}
\vspace{-.2in}
\caption{The bias, variance, and mean squared error (MSE) of the estimated gradient w.r.t. the number of samples for the PG estimator (left plot) and the RPG estimator (right plot). The shaded area representing a 95\% interval using bootstrapping techniques. The values of bias, variance, and MSE for the RPG estimator are significantly smaller than the values for the PG estimator, which is clear from the Y-axis ranges.}
\label{fig_lqg}
\end{figure}

We first briefly describe PPO, which is the basis of the new RPG algorithm.
PPO is simple to implement, closely related to on-policy learning, and achieves good performance on many tasks~\citep{schulman2017proximal}. The key idea behind PPO is to constrain the policy update by using a clipped surrogate objective $\E_{\pi_{\theta_{\text{old}}}}[l_t(\theta)]$, where $l_t(\theta) = \min(\rho_t(\theta) H_{t}, \text{clip}(\rho_t(\theta), 1-\eps, 1+\eps) H_{t})$, where $H_{t}$ is an estimator of the advantage function, and $\rho_t(\theta) = \pi_{\theta}(a_t | s_t) / \pi_{\theta_{\text{old}}}(a_t | s_t)$ is the importance sampling ratio. Then the gradients of $l_t(\theta)$ can be written as
\begin{equation} \label{eq_clip}
\nabla_{\theta} l_t(\theta)=\left\{
\begin{aligned}
& 0, & H_{t} > 0, \rho_t(\theta) > 1+\eps \\
& 0, & H_{t} < 0, \rho_t(\theta) < 1-\eps \\
& \nabla_{\theta} \rho_t(\theta) H_{t}, & \text{otherwise.}
\end{aligned}
\right.
\end{equation}

\begin{figure*}[htb]
\vspace{-.1in}
\begin{center}
  \subfigure[Peaks. RPG converges much faster than PPO in the whole training stage, when the variance is not so large.]{
    \includegraphics[width=\figwidthone]{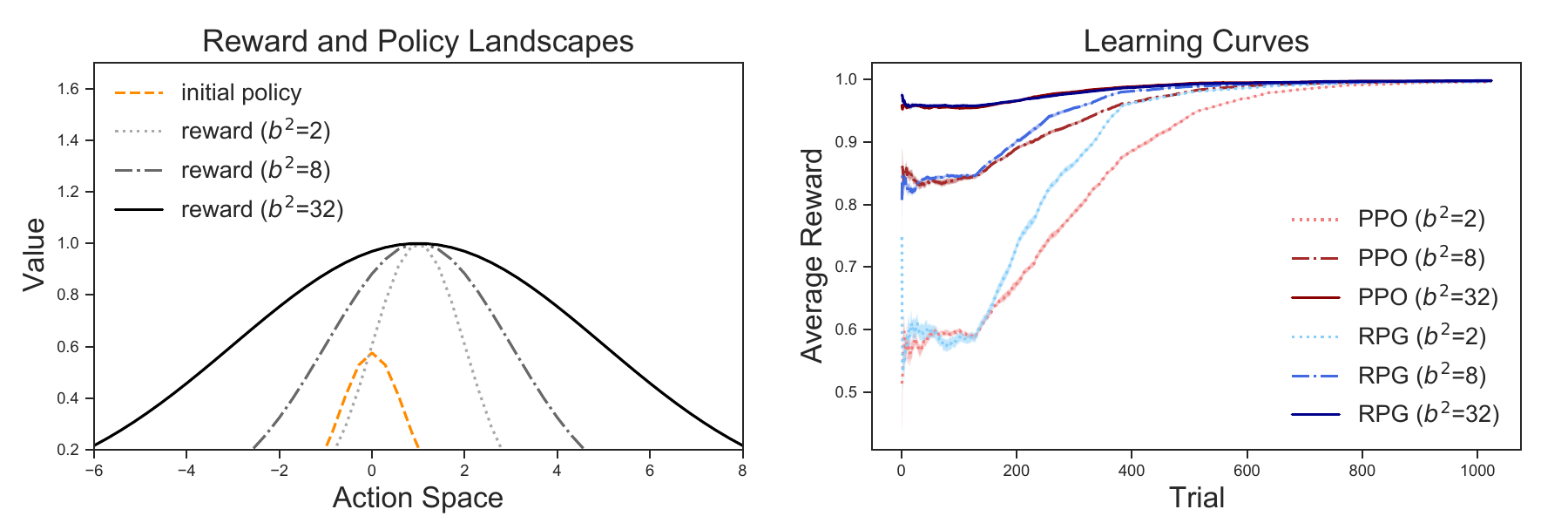}
    \label{fig_peak}
  }
\subfigure[Holes. RPG converges much faster than PPO at the early training stage, under various variance settings. Then it slows down at the later stage.]{
    \includegraphics[width=\figwidthone]{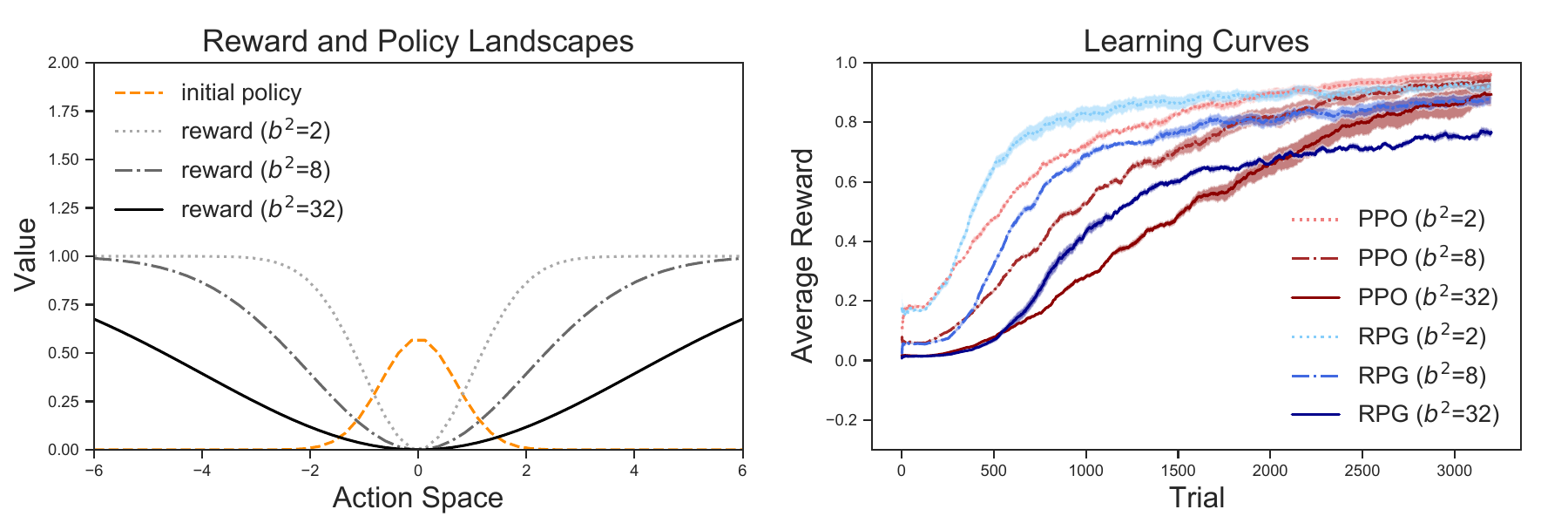}
    \label{fig_hole}
  }
\end{center}
\vspace{-.2in}
\caption{The reward landscapes of Peaks and Holes as well as learning curves for PPO and RPG during training. The initial Gaussian policies are also visualized. All results were averaged over 30 runs; the shaded areas represent standard errors.}
\end{figure*}

Inspired by Eq.~\ref{eq_clip}, we can use a modified ratio $\hat{\rho}_t(\theta)$,
\begin{equation} \label{eq_clip_new}
\hat{\rho}_t(\theta)=\left\{
\begin{aligned}
& 0, & H_{t} > 0, \rho_t(\theta) > 1+\eps \\
& 0, & H_{t} < 0, \rho_t(\theta) < 1-\eps \\
& \rho_t(\theta), & \text{otherwise.}
\end{aligned}
\right.
\end{equation}
Define $l_t(\theta) = \hat{\rho}_t(\theta) H_{t}$. It is easy to verify that this $\nabla_{\theta} l_t(\theta)$ is exactly the same as Eq.~\ref{eq_clip}. For a complete algorithm description of PPO, please check the appendix.

The RPG algorithm builds on PPO with two major modifications. First, we replace the original \LR{} in PPO with the RPG estimator. Second, in order to use reward gradients, we have a neural network to learn the reward function. Basically, RPG can be viewed as a version of PPO but using the RPG estimator to do gradient estimation. The detailed algorithm description for RPG is listed in Algorithm~\ref{algo_rpg}.

\begin{figure}[t]
\vspace{-.2in}
\includegraphics[width=\figwidthtwo]{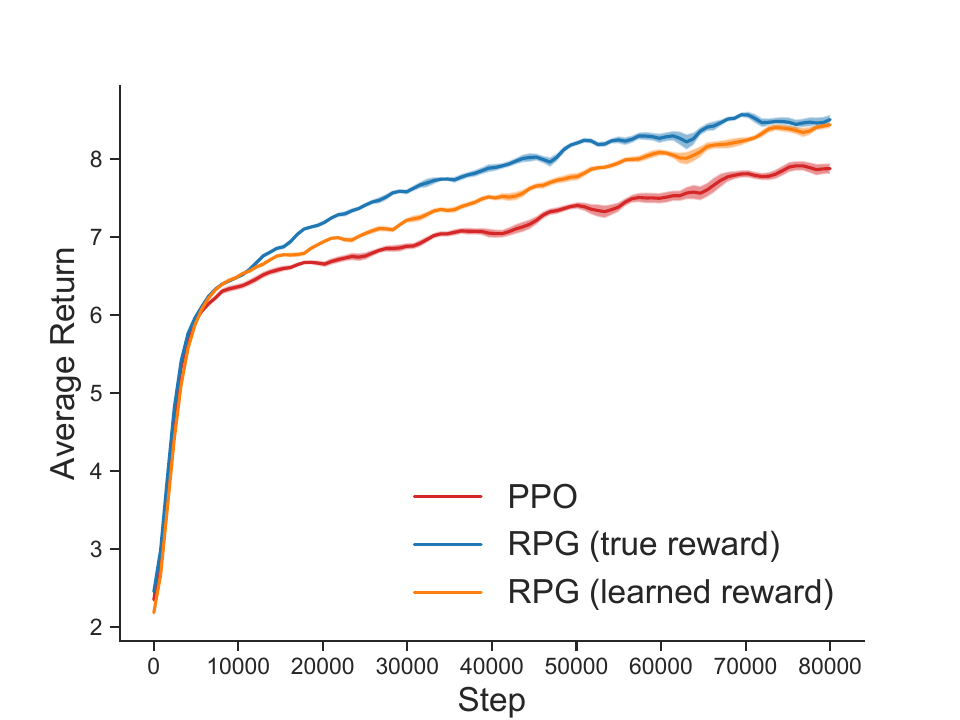}
\vspace{-.2in}
\caption{The learning curves for PPO and RPG during training on Mountain Climbing. The results were averaged over 20 runs, with the shaded area representing one standard error. In terms of convergence rate, RPG (true reward) > RPG (learned reward) > PPO.}
\label{fig_mountain}
\end{figure}

\begin{figure*}[!ht]
\vspace{-.2in}
\begin{center}
  \subfigure[HalfCheetah-v2]{\includegraphics[width=\figwidththree]{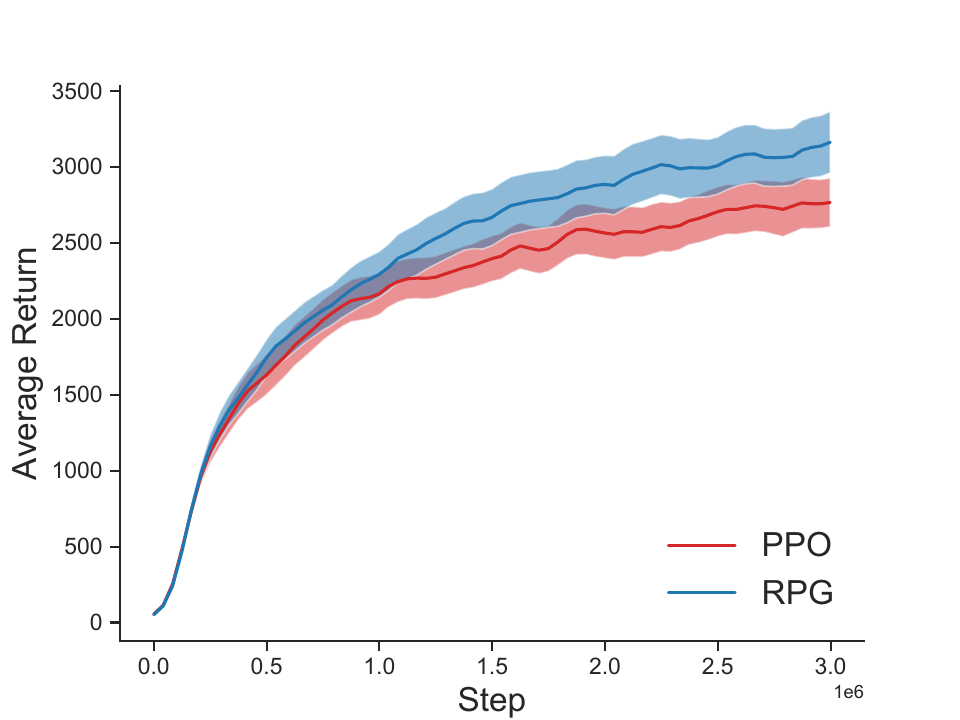}}\hspace{-0.7cm}\vspace{-0.1cm}
  \subfigure[Hopper-v2]{\includegraphics[width=\figwidththree]{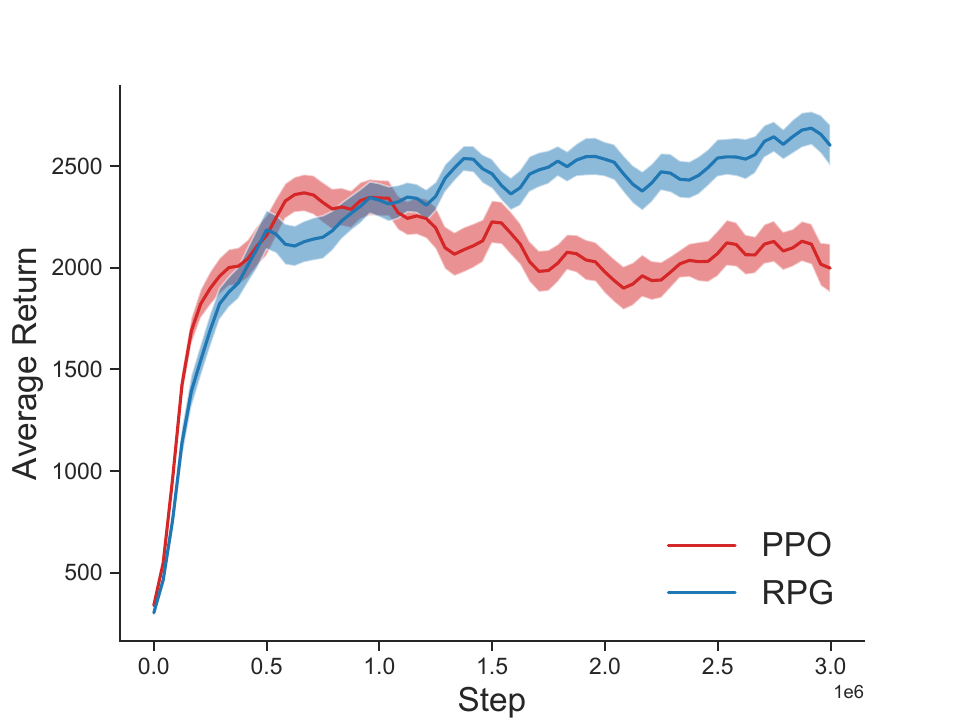}}\hspace{-0.7cm}\vspace{-0.1cm}
  \subfigure[Walker2d-v2]{\includegraphics[width=\figwidththree]{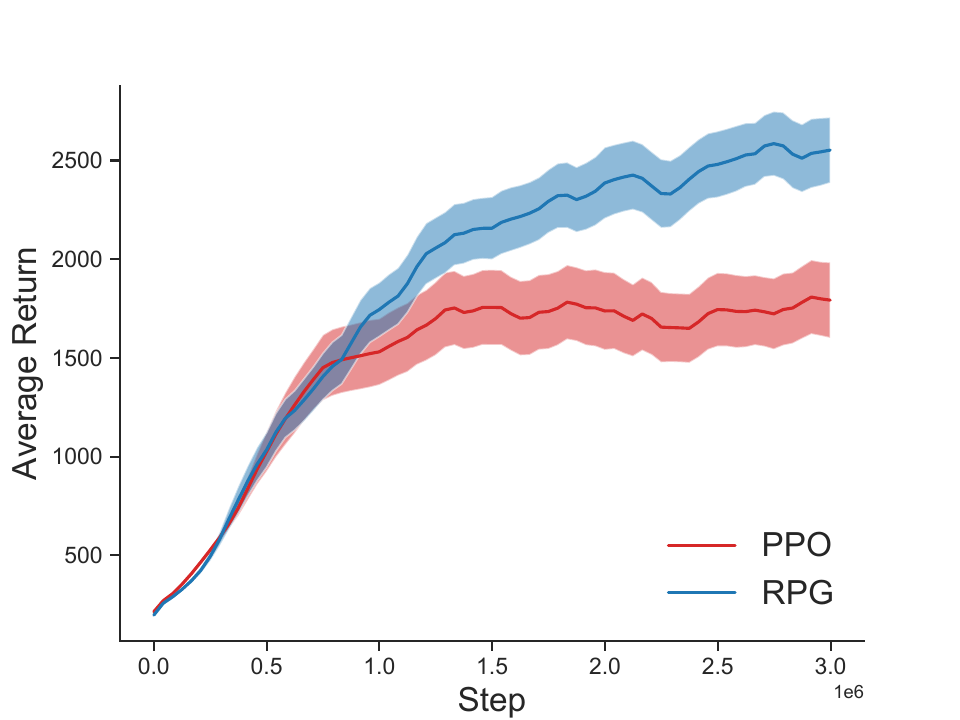}}\vspace{-0.1cm}
  \subfigure[Swimmer-v2]{\includegraphics[width=\figwidththree]{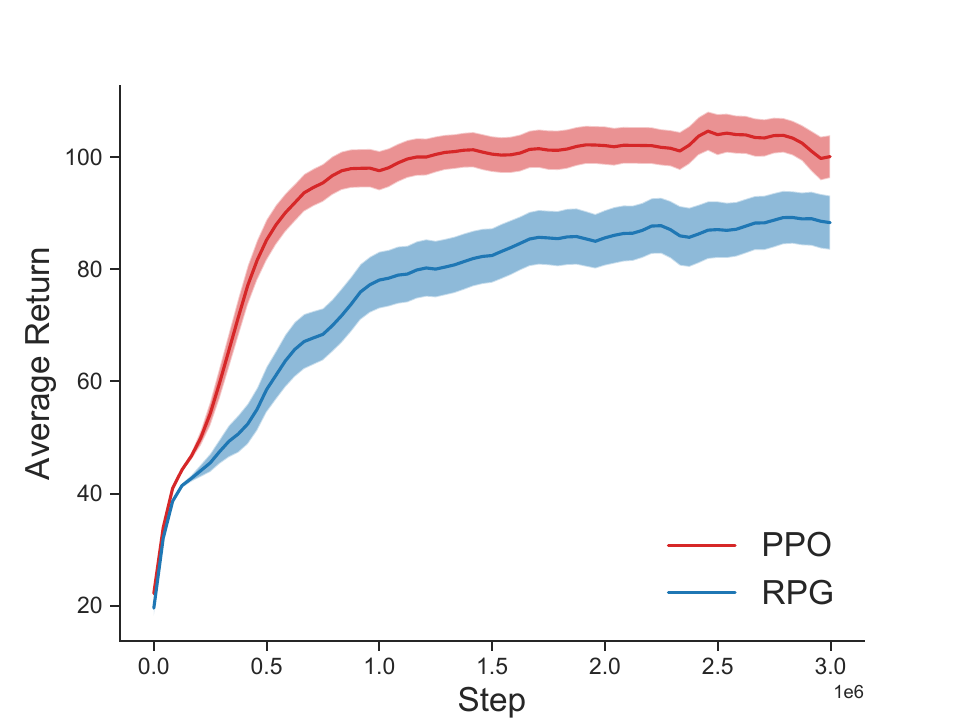}}\hspace{-0.7cm}
  \subfigure[Ant-v2]{\includegraphics[width=\figwidththree]{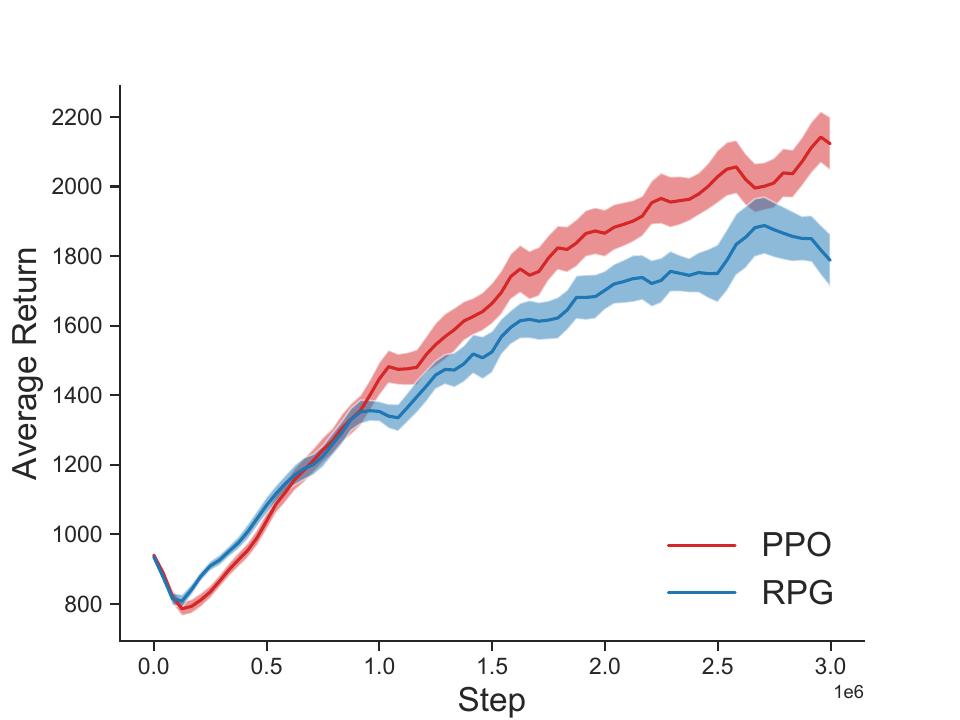}}\hspace{-0.7cm}
  \subfigure[Reacher-v2]{\includegraphics[width=\figwidththree]{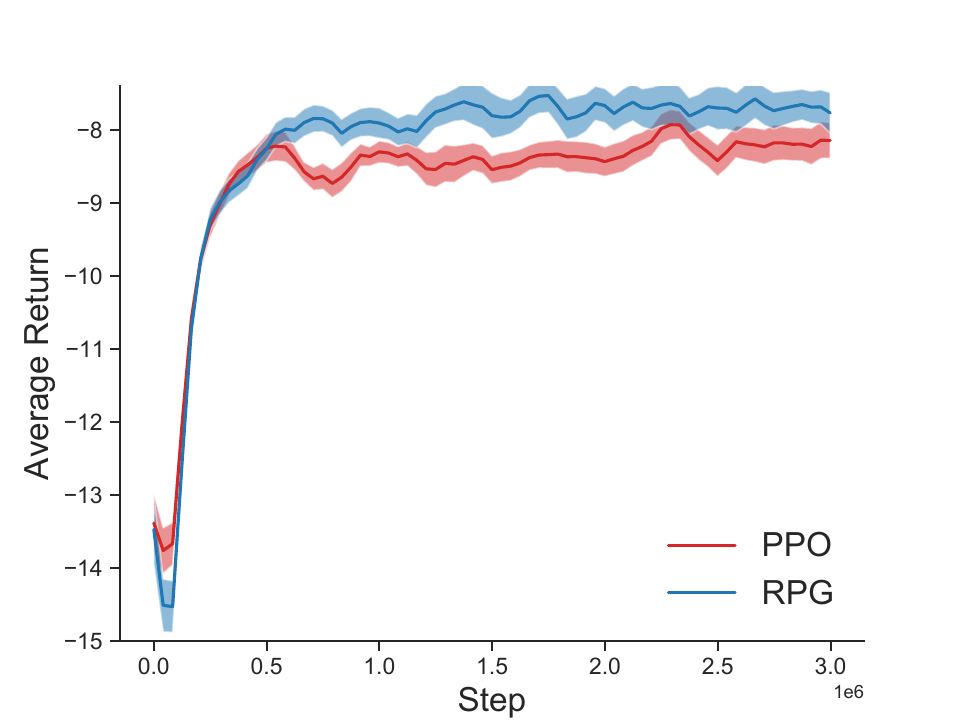}}
\end{center}
\vspace{-.2in}
\caption{Learning curves of evaluations on six benchmark tasks for PPO and RPG. The results were averaged over 30 runs, with the shaded area representing one standard error.}
\label{fig_mujoco}
\end{figure*}

\section{Experiments}

In this section, we first analyze the bias-variance trade-off for the RPG estimator on a Linear Quadratic Gaussian (LQG) control task. We then gain more understanding to the advantages and drawbacks of using reward gradients on two bandit tasks. 
Furthermore, we investigate the benefit of RPG when the reward function is known. Finally, we evaluate our algorithm on six MuJoCo control tasks~\citep{todorov2012mujoco} and one robot task compared to the baseline method.

\subsection*{A Bias-variance Analysis of the RPG Estimator}

Environments can have highly stochastic rewards~\citep{brockman2016openai}. Knowing the reward function in advance allows reducing the stochasticity of the gradient estimator. Furthermore, in most cases, reparameterization gradients exhibit lower variance w.r.t. likelihood ratio gradients~\citep{xu2019variance}. However, the variance of our estimator depends on many compounding factors (e.g. reward shapes and environment dynamics) that complicate a theoretical analysis. To compensate for this gap, we empirically investigate this aspect on a LQG problem.

The LQG control problem is one of the most classical control problems in control theory,  with linear dynamics, quadratic reward, and Gaussian noise. Using the Riccati equations, we can solve the LQG problem in closed form. This property makes the LQG task an ideal benchmark to do bias-variance analysis for different policy gradient estimators. Specifically, we consider the discrete-time LQG problem, defined as
\begin{align*}
& \max_{\theta} \sum_{t=0}^{\infty}  \gamma^t r_t \\
& \text{s.t. } \textbf{s}_{t+1} = A \textbf{s}_t + B \textbf{a}_t; \quad r_t = -{\textbf{s}_t}^\top Q \textbf{s}_t - {\textbf{a}_t}^\top Z \textbf{a}_t \\
& \quad\quad \textbf{a}_{t+1} = \Theta \textbf{s}_t + \Sigma \eps_t; \quad \eps_t \sim \gN(0, I), \\
\end{align*}
where $A,B,Q,Z,\Sigma$, and $\Theta$ are diagonal matrices and $\Theta = \text{diag}(\theta)$. Our policy is Gaussian: $\textbf{a} \sim \gN(\Theta \textbf{s}; \Sigma)$, where $\theta \in \sR^2$.

Given an LQG task, we study how the bias, the variance, and the mean squared error (MSE) of the estimated gradient vary w.r.t. the number of samples for both the policy gradient (PG) estimator and the RPG estimator. A similar study can also be found in~\cite{tosatto2021batch}. Note that we do not subtract a baseline term in either estimator as in Eq.~\ref{baseline}. In our experiment, we assume that the true reward function and the true value function are provided to both gradient estimators. Once the reward function is given, the RPG estimator is able to use not only the reward signals but also reward gradients. However, the PG estimator can only use the reward signals since it is not designed to use reward gradients. For further details, please refer to the appendix.
In Figure~\ref{fig_lqg}, the left and right plots show the values of the PG estimator and the RPG estimator, respectively. Clearly, both the bias and the variance of the RPG estimator are significantly smaller than the ones of the PG estimator, which can be noticed from the Y-axis ranges of the two plots. 
This result indicates that by incorporating reward gradients, the RPG estimator is able to reduce the bias and the variance of gradient estimation substantially.

\subsection*{Benefits and Drawbacks of Reward Gradients}

To further understand the role of reward gradients, we design two kinds of bandit tasks---\textit{Peaks} and \textit{Holes}---with continuous action spaces $\real$. The reward functions of Peaks and Holes are defined as $r(a)=\exp\left(-\frac{(a-1)^2}{b^2}\right)$ and $r(a)=1-\exp\left(-\frac{a^2}{b^2}\right)$, as shown in Figure~\ref{fig_peak} and Figure~\ref{fig_hole}, respectively. In our experiments, a small ($0.01\times$) Gaussian noise is added to all reward signals and $b^2$ is chosen from $[2, 8, 32]$.

The initial policy distribution is a Gaussian distribution $\gN(0, 0.69)$. For RPG, we use a neural network to approximate the true reward function, denoted by $\hat{r}_w(a)$. In bandit cases, there are no value functions and we have the following single-sample gradients for PPO and RPG, without considering normalization for simplicity:
\begin{align} \label{eq_bandit_grad}
&\nabla_{\theta} l_t^{\text{PPO}}(\theta) = \hat{\rho}_t(\theta) R_t \nabla_{\theta} \log \pi_{\theta}(A_t)
\text{ and } \nonumber \\
&\nabla_{\theta} l_t^{\text{RPG}}(\theta) = \hat{\rho}_t(\theta) \nabla_{\theta} f_{\theta}(\eps_t) \nabla_{a} \hat{r}_w(a)|_{a=A_t},
\end{align}
where $\theta = [\mu, \sigma]$. We test PPO and RPG on these bandits over 30 runs. The gradient is clipped by 1.

The learning curves of PPO and RPG are visualized in Figure~\ref{fig_peak} and \ref{fig_hole}. We observe that as $b^2$ increases, the learning speed decreases for both PPO and RPG, on Peaks and Holes. This is because the magnitude of reward's gradient decreases as $b^2$ increases which reduces $\nabla_{\theta} l_t(\theta)$ and slows down the learning update, as suggested by Eq.~\ref{eq_bandit_grad}. 

Furthermore, RPG converges much faster than PPO on Peaks in the whole training period when $b^2$ is not so large. This shows that RPG is able to find the optimal action quicker than PPO, with the help of a relatively large reward's gradient. On Holes, RPG escapes from the sub-optimal action faster than PPO at the early training stage, but it slows down later and performs worse than PPO in the end. This is not surprising since the reward function is still sharp at the early training stage (when $|a|$ is small) but tends to become flat later (i.e. $|\nabla_{a} \hat{r}_w(a)|$ is smaller.) which slows down the learning process of RPG.

Therefore, we conclude that in bandits the reward's gradient accelerates learning when it is relatively large but hurts learning when the reward function is too flat. 
Reward and action-value gradients are identical in bandits; to compensate that, and we conduct the rest of the studies using simple and complex MDPs.

\subsection*{The Benefit of Knowing the Reward Function}

To explore the role of reward gradients in MDP settings, we design a simple MDP---\textit{Mountain Climbing}---with continuous state and action spaces. Specifically, $\States=[-8,8]^2$, $\Actions=[-1,1]^2$, $S_0=(0,0), R_{t+1} = r(S_t, A_t)$, and $S_{t+1} = S_t + A_t + \eps$ where $r(s,a) = \exp(-||s+a-\nu||_2^2)$, $\eps \sim \gU(-0.005,0.005)$, and $\nu=(1,-1)$. Every episode ends in 10 steps. 
We test PPO (LR gradient) and RPG (LR gradient + reward gradient) on this MDP for $80,000$ steps. We set the number of epochs in PPO and RPG to 1 and the batch and  mini-batch size to 40. The gradient is clipped by 0.5.

The average returns over 20 runs for each algorithm during training are shown in Figure~\ref{fig_mountain}. By utilizing reward gradients, RPG (LR gradient + reward gradient) outperforms PPO (LR gradient) significantly. Moreover, the reward gradients of the true reward function are more helpful to accelerate learning than the reward gradients of a learned one, probably due to a higher accuracy of gradient estimation; although two versions of RPG reach a similar performance in the end.

\subsection*{Evaluation on Simulated Benchmark Tasks}

To further evaluate our algorithm, we measure its performance on six MuJoCo control tasks~\citep{todorov2014convex} through OpenAI Gym~\citep{brockman2016openai}.
Our PPO implementation is based on \citet{deeprl} and \citet{SpinningUp2018}. Then we implement RPG based on this version of PPO. The appendix contains all hyper-parameters which are largely borrowed from some popular implementations. The code is also submitted as a supplementary file. For the simulated tasks, each algorithm was trained on every task for 3 million steps. For every 5 epochs, we evaluated the agent's test performance using a deterministic policy for one episode. Our results were reported by averaging over 30 runs with different random seeds. The learning curves during evaluation are presented in Figure~\ref{fig_mujoco}. 
Overall, RPG outperforms PPO significantly on three tasks -- HalfCheetah, Hopper, and Walker2d. It is slightly better on Reacher and worse on Ant, compared to PPO. On Swimmer, however, PPO has a clear advantage.

\subsection*{Evaluation on a Real-Robot Task}
We include results on a real-robot task to showcase RPG's ability to effectively learn in a real-world scenario.
We use the Real-Robot Reacher task based on a UR robotic arm in which an agent must move the fingertip of a robot arm as close and as fast as possible to a randomly selected target by rotating the base and elbow joints.
Each episode is 4 seconds long followed by a reset, bringing the fingertip to a start position.
The appendix contains the full description of the task.
RPG runs on Real-Robot Reacher for 90,000 time steps at 40ms action-cycle time, resulting in an hour of robot-experience time excluding resets.
The undiscounted episodic returns were stored during training without measuring a separate test performance.

\begin{figure}[t]
\vspace{-.2in}
\includegraphics[width=\figwidthtwo]{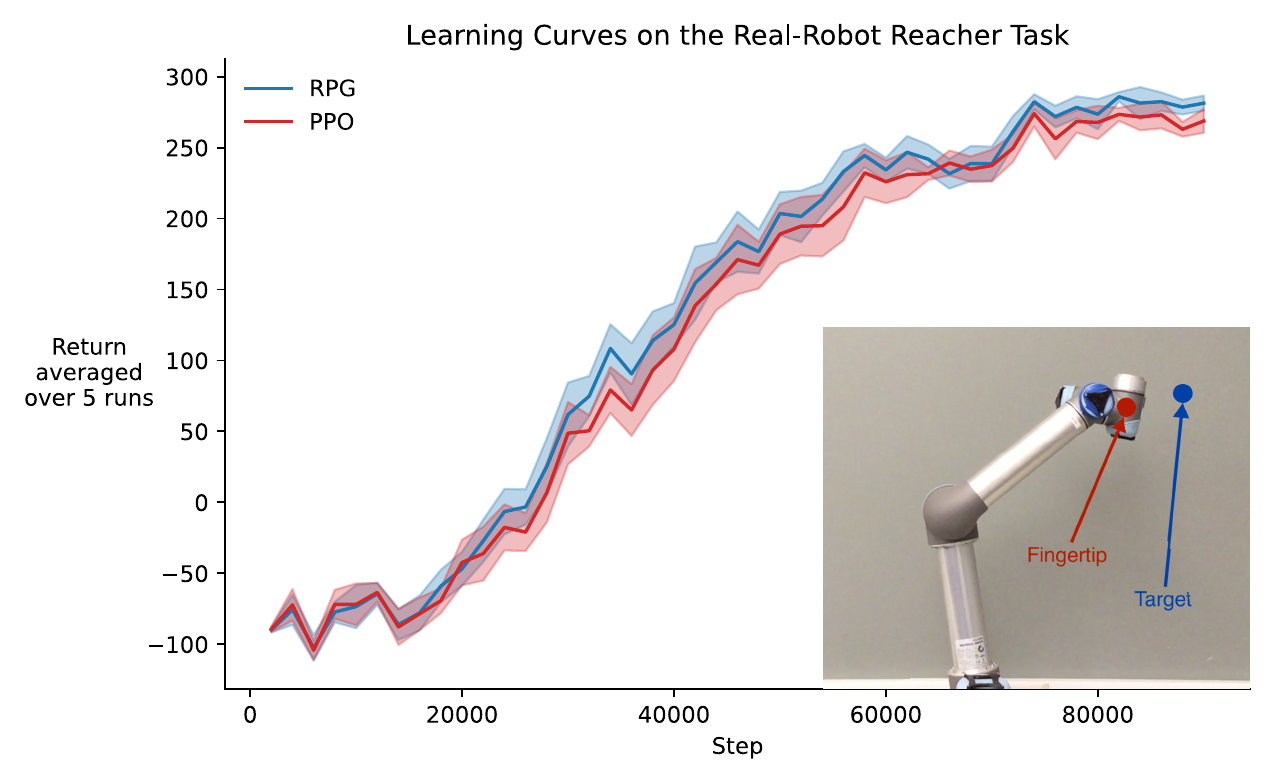}
\vspace{-.2in}
\caption{The learning curves for RPG and PPO on the Real-Robot Reacher task. The results were averaged over 5 runs, with the shaded area representing one standard error. The bottom-right image depicts the task setup. RPG is as effective as PPO on this task.}
\label{fig:ur5}
\end{figure}

Figure~\ref{fig:ur5} shows the resulting learning curves averaged over five independent runs.
RPG continues to increase performance throughout learning.
For comparison, we also included the performance of PPO, which is known to learn an effective reaching behavior on this task, specifically at this performance level of about 300 average return~\citep{mahmood2018benchmarking,farrahi2020making}.
Both RPG and PPO achieved the same performance on this task, which is not surprising as we did not perform a hyper-parameter search.
Instead, our results confirm that RPG is at least as effective as PPO on a well-studied real-robot task.

\section{Discussion}

Our work focuses more on understanding the properties of the RPG estimator by conducting a series of analysis on simple environments. 
Moreover, the RPG theorem only provides a new way to estimate policy gradient; the implementation of actual algorithms can vary. For example, the RPG version of the na\"ive actor-critic would be a straight-forward implementation. Our current implementation builds on PPO, and it benefits from PPO's techniques as well. To further explore and exploit the advantage of the RPG estimator, we may develop more advanced implementations by combining some modern techniques in the future, such as entropy regularization~\citep{haarnoja2018soft}, parallel training~\citep{mnih2016asynchronous}, separating training phases for policy and value functions~\citep{cobbe2021phasic}, Retrace~\citep{munos2016safe}, and V-trace~\citep{espeholt2018impala}, etc. Our current implementation is just one approach; exploring more possibilities are among the potential future works resulting from this work.

\section{Conclusion}

In this paper, we introduced a novel strategy to compute the policy gradient which uses reward gradients without a model. Based on this strategy, we developed---RPG---a new on-policy policy gradient algorithm. We showed that our method of using reward gradients is beneficial over the PG estimator in terms of the bias-variance trade-off and sample efficiency. Experiments showed that RPG generally outperformed PPO on several simulation tasks.
RPG is limited to the same class of tasks where reparameterization applies and, thus, not directly applicable to tasks with discrete actions. However, combined with the Gumbel-Softmax technique~\citep{jang2016categorical}, it is possible to apply RPG on discrete control tasks as well.

\subsubsection*{Acknowledgements}
We gratefully acknowledge funding from the Canada CIFAR AI Chairs program, the Reinforcement Learning and Artificial Intelligence (RLAI) laboratory, the Alberta Machine Intelligence Institute (Amii), and the Natural Sciences and Engineering Research Council (NSERC) of Canada.
We also thankfully acknowledge the donation of the UR5 arm from the Ocado Group.

\bibliography{reference}

\clearpage
\appendix
\thispagestyle{empty}
\onecolumn \makesupplementtitle
\input{appendix.tex}

\end{document}

%% file: math_commands.tex
\usepackage{amsmath,amsfonts,bm}

\def\eqref#1{equation~\ref{#1}}

\def\1{\bm{1}}

\def\eps{{\epsilon}}

\DeclareMathAlphabet{\mathsfit}{\encodingdefault}{\sfdefault}{m}{sl}
\SetMathAlphabet{\mathsfit}{bold}{\encodingdefault}{\sfdefault}{bx}{n}

\def\gH{{\mathcal{H}}}

\def\gN{{\mathcal{N}}}

\def\gU{{\mathcal{U}}}

\def\sR{{\mathbb{R}}}

\newcommand{\real}{\mathbb{R}}
\newcommand{\E}{\mathbb{E}}

\newcommand{\de}{\mathrm{d}}

\DeclareMathOperator*{\argmin}{arg\,min}

%% file: figures/lqg.tex
\begin{tikzpicture}

\definecolor{color0}{rgb}{0.12156862745098,0.466666666666667,0.705882352941177}
\definecolor{color1}{rgb}{1,0.498039215686275,0.0549019607843137}
\definecolor{color2}{rgb}{0.172549019607843,0.627450980392157,0.172549019607843}

\begin{groupplot}[group style={group size=2 by 1,
            horizontal sep=15pt}]
\nextgroupplot[
title={\footnotesize (a) PG},
width=0.59 \columnwidth,
height=0.56 \columnwidth,
legend cell align={left},
legend style={fill opacity=0.8, draw opacity=1, text opacity=1, draw=white!80!black},
tick align=outside,
tick pos=left,
x grid style={white!69.0196078431373!black},
xmin=5.5, xmax=104.5,
scaled y ticks=false,
yticklabels = {$0$, $0$, $2 \times 10^4$, $4 \times 10^4$, $6 \times 10^4$, $8 \times 10^4$, $1 \times 10^5$},
xtick style={color=black},
xtick={0,20,40,60,80,100,120},
yticklabel  style={font=\tiny,  xshift=3pt},
xticklabel  style={font=\tiny, yshift=3pt},
	grid=both,
	grid style={line width=.1pt, draw=gray!10},
xlabel={\scriptsize \# Samples},
xticklabels={
  \(\displaystyle {0}\),
  \(\displaystyle {20}\),
  \(\displaystyle {40}\),
  \(\displaystyle {60}\),
  \(\displaystyle {80}\),
  \(\displaystyle {100}\),
  \(\displaystyle {120}\)
},
y grid style={white!69.0196078431373!black},
ymin=-4840.12902322065, ymax=100688.948615721,
ytick style={color=black},
ytick={-20000,0,20000,40000,60000,80000,100000,120000},
]
\path [fill=color0, fill opacity=0.5]
(axis cs:10,-43.3527669051091)
--(axis cs:10,66.0588155891591)
--(axis cs:25,69.4185288652231)
--(axis cs:50,383.343919395545)
--(axis cs:75,56.332766244758)
--(axis cs:100,97.6284432726448)
--(axis cs:100,61.8654430105376)
--(axis cs:100,61.8654430105376)
--(axis cs:75,16.2554047585117)
--(axis cs:50,334.54195489987)
--(axis cs:25,-2.52342857967426)
--(axis cs:10,-43.3527669051091)
--cycle;

\path [fill=color1, fill opacity=0.5]
(axis cs:10,95892.1723594057)
--(axis cs:10,64668.6312307341)
--(axis cs:25,27934.5251646921)
--(axis cs:50,12854.6498297774)
--(axis cs:75,8692.9133420804)
--(axis cs:100,6902.72575612619)
--(axis cs:100,10235.5246331792)
--(axis cs:100,10235.5246331792)
--(axis cs:75,12890.0570282673)
--(axis cs:50,19061.1780957409)
--(axis cs:25,41421.9730786222)
--(axis cs:10,95892.1723594057)
--cycle;

\path [fill=color2, fill opacity=0.5]
(axis cs:10,62302.5216714864)
--(axis cs:10,92898.8261207598)
--(axis cs:25,39937.8130064779)
--(axis cs:50,18822.4435765293)
--(axis cs:75,12384.8849214486)
--(axis cs:100,9977.92748078314)
--(axis cs:100,6745.32516733528)
--(axis cs:100,6745.32516733528)
--(axis cs:75,8547.19044676365)
--(axis cs:50,12741.4191293576)
--(axis cs:25,27160.6797990662)
--(axis cs:10,62302.5216714864)
--cycle;

\addplot [semithick, color0]
table {%
10 11.3530244827271
25 33.4475517272949
50 358.942932128906
75 36.2940864562988
100 79.746940612793
};
\addplot [semithick, color1]
table {%
10 77979.21875
25 33684.21875
50 15500.4912109375
75 10482.154296875
100 8323.4970703125
};
\addplot [semithick, color2]
table {%
10 77600.671875
25 33549.24609375
50 15781.931640625
75 10466.0380859375
100 8361.6259765625
};

\nextgroupplot[
title={\footnotesize (b) RPG},
width=0.59 \columnwidth,
height=0.56 \columnwidth,
legend cell align={left},
legend style={fill opacity=0.8, draw opacity=1, text opacity=1, draw=white!80!black},
tick align=outside,
tick pos=left,
x grid style={white!69.0196078431373!black},
xmin=5.5, xmax=104.5,
xtick style={color=black},
xtick={0,20,40,60,80,100,120},
yticklabel  style={font=\tiny, xshift=3pt},
xticklabel  style={font=\tiny, yshift=3pt},
	grid=both,
	grid style={line width=.1pt, draw=gray!10},
xlabel={\scriptsize \# Samples},
xticklabels={
  \(\displaystyle {0}\),
  \(\displaystyle {20}\),
  \(\displaystyle {40}\),
  \(\displaystyle {60}\),
  \(\displaystyle {80}\),
  \(\displaystyle {100}\),
  \(\displaystyle {120}\)
},
legend style={font=\footnotesize},
y grid style={white!69.0196078431373!black},
ymin=-1.18812358402525, ymax=12.1299363353144,
ytick style={color=black},
ytick={-2,0,2,4,6,8,10,12,14},
yticklabels={
  \(\displaystyle {−2}\),
  \(\displaystyle {0}\),
  \(\displaystyle {2}\),
  \(\displaystyle {4}\),
  \(\displaystyle {6}\),
  \(\displaystyle {8}\),
  \(\displaystyle {10}\),
  \(\displaystyle {12}\),
  \(\displaystyle {14}\)
}
]
\path [fill=color0, fill opacity=0.5]
(axis cs:10,-0.582757224055272)
--(axis cs:10,0.617054212106791)
--(axis cs:25,0.379416152068239)
--(axis cs:50,0.258782151591049)
--(axis cs:75,0.201181309106154)
--(axis cs:100,0.17203654218544)
--(axis cs:100,-0.169632494527324)
--(axis cs:100,-0.169632494527324)
--(axis cs:75,-0.200925169518146)
--(axis cs:50,-0.233336415602471)
--(axis cs:25,-0.337793599287389)
--(axis cs:10,-0.582757224055272)
--cycle;

\path [fill=color1, fill opacity=0.5]
(axis cs:10,11.5245699753444)
--(axis cs:10,7.7720438226701)
--(axis cs:25,2.77993901797258)
--(axis cs:50,1.30902306781606)
--(axis cs:75,0.873318646814248)
--(axis cs:100,0.630277410755229)
--(axis cs:100,0.934590217175596)
--(axis cs:100,0.934590217175596)
--(axis cs:75,1.29497749699076)
--(axis cs:50,1.94105029366695)
--(axis cs:25,4.12215917341656)
--(axis cs:10,11.5245699753444)
--cycle;

\path [fill=color2, fill opacity=0.5]
(axis cs:10,7.55660300181105)
--(axis cs:10,11.1274656146247)
--(axis cs:25,4.02939797703815)
--(axis cs:50,1.96317209392054)
--(axis cs:75,1.25120350164974)
--(axis cs:100,0.902813880929296)
--(axis cs:100,0.61200183620858)
--(axis cs:100,0.61200183620858)
--(axis cs:75,0.844665284175226)
--(axis cs:50,1.20340123064993)
--(axis cs:25,2.68295739827222)
--(axis cs:10,7.55660300181105)
--cycle;

\addplot [semithick, color0]
table {%
10 0.017148494720459
25 0.0208113193511963
75 0.000128030776977539
100 0.00120198726654053
};
\addlegendentry{Bias\textsuperscript{2}}
\addplot [semithick, color1]
table {%
10 9.37174415588379
25 3.35212707519531
50 1.57845604419708
75 1.05307173728943
100 0.760005950927734
};
\addlegendentry{Variance}
\addplot [semithick, color2]
table {%
10 9.34203433990479
25 3.35617780685425
50 1.58328664302826
75 1.04793441295624
100 0.757407903671265
};
\addlegendentry{MSE}
\end{groupplot}

\end{tikzpicture}

%% file: appendix.tex
\section{Theorems and Proofs}

\subsection{Proof of Reward Policy Gradient Theorem}

Let $d^{\pi_{\theta}}(s') = \int \sum_{t=0}^{\infty} \gamma^t p_0(s) p(s \to s', t, \pi_{\theta}) \de s$ be the (discounted) stationary state distribution for policy $\pi_{\theta}$; and $p(s \to s', t, \pi_{\theta})$  the transition probability from $s$ to $s'$ with $t$ steps under policy $\pi_{\theta}$.

\begin{reptheorem}{thm_rpg}[Reward Policy Gradient Theorem]
Suppose that the MDP satisfies Assumption~\ref{assum1} and \ref{assum2}, then
\begin{align*}
\nabla_{\theta} J(\theta) = \int d^{\pi_{\theta}}(s) \pi_{\theta}(a|s) p(s'|s,a) \left[\nabla_{\theta} r(s,f_{\theta}(\eps; s))|_{\epsilon=g_{\theta}(a;s)} + \gamma v_{\pi_{\theta}}(s') \nabla_{\theta} \log{\pi_{\theta}(a | s)} \right] \de s \de a \de s'.
\end{align*}
\end{reptheorem}

\begin{proof}

By the policy gradient theorem, we have
\begin{align*}
\nabla_{\theta} J(\theta)
= \int d^{\pi_{\theta}}(s) \pi_{\theta}(a|s) q_{\pi_{\theta}}(s,a) \nabla_{\theta} \log{\pi_{\theta}(a|s)} \de a \de s.
\end{align*}

Next, we split the action-value function $q_{\pi_{\theta}}(s,a)$ into two parts:
\begin{align*}
\nabla_{\theta} J(\theta)
=& \int d^{\pi_{\theta}}(s) \pi_{\theta}(a|s) \left[r(s,a) + \gamma \int p(s' | s,a) v_{\pi_{\theta}}(s') \de s' \right] \nabla_{\theta} \log{\pi_{\theta}(a|s)} \de a \de s \\
=& \int d^{\pi_{\theta}}(s) \pi_{\theta}(a|s) r(s,a) \nabla_{\theta} \log{\pi_{\theta}(a|s)} \de s \de a + \gamma \int d^{\pi_{\theta}}(s) \pi_{\theta}(a|s) p(s'|s,a) v_{\pi_{\theta}}(s') \nabla_{\theta} \log{\pi_{\theta}(a|s)} \de s \de a \de s' \\
=& \int d^{\pi_{\theta}}(s) r(s,a) \nabla_{\theta} \pi_{\theta}(a|s) \de s \de a + \gamma \int d^{\pi_{\theta}}(s) \pi_{\theta}(a|s) p(s'|s,a) v_{\pi_{\theta}}(s') \nabla_{\theta} \log{\pi_{\theta}(a|s)} \de s \de a \de s' \\
=& \int d^{\pi_{\theta}}(s) \nabla_{\theta} \left( \int \pi_{\theta}(a|s) r(s,a) \de a \right) \de s + \gamma \int d^{\pi_{\theta}}(s) \pi_{\theta}(a|s) p(s'|s,a) v_{\pi_{\theta}}(s') \nabla_{\theta} \log{\pi_{\theta}(a|s)} \de s \de a \de s'.
\end{align*}

Now, we apply the reparameterization technique to the first part,
\begin{align*}
\nabla_{\theta} J(\theta)
=& \int d^{\pi_{\theta}}(s) \nabla_{\theta} \left( \int \pi_{\theta}(a|s) r(s,a) \de a \right) \de s + \gamma \int d^{\pi_{\theta}}(s) \pi_{\theta}(a|s) p(s'|s,a) v_{\pi_{\theta}}(s') \nabla_{\theta} \log{\pi_{\theta}(a|s)} \de s \de a \de s' \\
=& \int d^{\pi_{\theta}}(s) \nabla_{\theta} \left( \int p(\eps) r(s,f_{\theta}(\eps; s)) \de \eps \right) \de s + \gamma \int d^{\pi_{\theta}}(s) \pi_{\theta}(a|s) p(s'|s,a) v_{\pi_{\theta}}(s') \nabla_{\theta} \log{\pi_{\theta}(a|s)} \de s \de a \de s' \\
=& \int d^{\pi_{\theta}}(s) \left( \int p(\eps) \nabla_{\theta} r(s,f_{\theta}(\eps; s)) \de \eps \right) \de s + \gamma \int d^{\pi_{\theta}}(s) \pi_{\theta}(a|s) p(s'|s,a) v_{\pi_{\theta}}(s') \nabla_{\theta} \log{\pi_{\theta}(a|s)} \de s \de a \de s'.
\end{align*}

We then apply the reverse operation of reparameterization to the first part,
\begin{align*}
&\nabla_{\theta} J(\theta) \\
=& \int d^{\pi_{\theta}}(s) \left( \int p(\eps) \nabla_{\theta} r(s,f_{\theta}(\eps; s)) \de \eps \right) \de s + \gamma \int d^{\pi_{\theta}}(s) \pi_{\theta}(a|s) p(s'|s,a) v_{\pi_{\theta}}(s') \nabla_{\theta} \log{\pi_{\theta}(a|s)} \de s \de a \de s' \\
=& \int d^{\pi_{\theta}}(s) \left( \int \pi_{\theta}(a|s) \nabla_{\theta} r(s,f_{\theta}(\eps; s))|_{\epsilon=g_{\theta}(a;s)} \de a \right) \de s + \gamma \int d^{\pi_{\theta}}(s) \pi_{\theta}(a|s) p(s'|s,a) v_{\pi_{\theta}}(s') \nabla_{\theta} \log{\pi_{\theta}(a|s)} \de s \de a \de s' \\
=& \int d^{\pi_{\theta}}(s) \pi_{\theta}(a|s) p(s'|s,a) \left[\nabla_{\theta} r(s,f_{\theta}(\eps; s))|_{\epsilon=g_{\theta}(a;s)} + \gamma v_{\pi_{\theta}}(s') \nabla_{\theta} \log{\pi_{\theta}(a | s)} \right] \de s \de a \de s'.
\end{align*}

\end{proof}

\subsection{Proof of Reward Policy Gradient with Function Approximation}

\begin{reptheorem}{thm_rpg_appro}[Reward Policy Gradient with Function Approximation]
Consider a parametric approximation of the reward function $\hat{r}_\omega(s, a)$ and a parametric approximation of the value function $\hat{v}_\phi(s)$ such that
\begin{enumerate}
\item $\nabla_a \hat{r}_\omega(s, a) = \nabla_\theta^\intercal f_\theta(\epsilon; s)|_{\epsilon = g_{\theta}(a; s)} \omega$,
\item $\int d^{\pi_{\theta}}(s) \pi_{\theta}(a|s) p(s'|s,a) \nabla_{\phi} \hat{v}_{\phi}(s') \de s \de a = \int d^{\pi_{\theta}}(s) \pi_{\theta}(a|s) p(s'|s,a) \nabla_{\theta} \log \pi_{\theta}(a|s) \de s \de a$,
\end{enumerate}
where $\phi = \argmin_{\phi} \underset{\substack{s \sim d^{\pi_{\theta}} \\ a \sim \pi_\theta \\ s' \sim p(\cdot|s, a)}}{\E}\left[(\hat{v}_\phi(s') - v(s'))^2\right]$ \\
and $\omega = \argmin_{\omega} \underset{\substack{s \sim d^{\pi_{\theta}} \\ a \sim \pi_\theta }}{\E} \left[(\nabla_a \hat{r}_\omega(s, a) - \nabla_a r(s, a))^\intercal (\nabla_a \hat{r}_\omega(s, a) - \nabla_a r(s,a)) \right]$. \\
Then, 
$\nabla_\theta J(\theta)
= \underset{\substack{s \sim d^{\pi_{\theta}} \\ a \sim \pi_\theta \\ s' \sim p(\cdot|s,a)}}{\E} \left[\nabla_\theta\hat{r}_{\omega}(s, f_\theta(\epsilon; s))|_{\epsilon = g_\theta(a; s)} + \gamma \hat{v}_{\phi}(s')\nabla_\theta \log \pi_{\theta}(a | s) \right]$.
\end{reptheorem}

\begin{proof}
By the definitions of $\phi$ and $\omega$, we have
\begin{align*}
& \omega  = \argmin_{\omega}\underset{\substack{s \sim d^{\pi_{\theta}} \\ a \sim \pi_\theta }}{\E}\left[(\nabla_a \hat{r}_\omega(s, a) - \nabla_a r(s, a))^\intercal(\nabla_a \hat{r}_\omega(s, a) - \nabla_a r(s, a)) \right] \\
\implies & \underset{\substack{s \sim d^{\pi_{\theta}} \\ a \sim \pi_\theta }}{\E}\left[(\nabla_a \hat{r}_\omega(s, a) - \nabla_a r(s, a))\nabla_\omega \nabla_a \hat{r}_\omega(s, a) \right] = 0 \\
\implies & \underset{\substack{s \sim d^{\pi_{\theta}} \\ a \sim \pi_\theta }}{\E}\left[(\nabla_a \hat{r}_\omega(s, a) - \nabla_a r(s, a))\nabla_\theta   f_\theta(\epsilon; s)|_{\epsilon = g(a; s)} \right] = 0 \\
\implies & \underset{\substack{s \sim d^{\pi_{\theta}} \\ a \sim \pi_\theta }}{\E}\left[\nabla_a \hat{r}_\omega(s, a)\nabla_\theta f_\theta(\epsilon; s)|_{\epsilon = g(a; s)} \right] =  \underset{\substack{s \sim d^{\pi_{\theta}} \\ a \sim \pi_\theta }}{\E}\left[\nabla_a r(s, a)\nabla_\theta f_\theta(\epsilon; s)|_{\epsilon = g(a; s)} \right] \\
\implies & \underset{\substack{s \sim d^{\pi_{\theta}} \\ a \sim \pi_\theta }}{\E}\left[\nabla_\theta\hat{r}_{\omega}(s, f_\theta(\epsilon; s))|_{\epsilon = g_\theta(a; s)} \right] =  \underset{\substack{s \sim d^{\pi_{\theta}} \\ a \sim \pi_\theta }}{\E}\left[\nabla_\theta r(s, f_\theta(\epsilon; s))|_{\epsilon=g_\theta(a; s)} \right],
\end{align*}
and
\begin{align*}
& \phi = \argmin_\phi \underset{\substack{s \sim d^{\pi_{\theta}} \\ a \sim \pi_\theta \\
s' \sim p(\cdot | s, a)}}{\E}\left[(\hat{v}_{\phi}(s') - v(s') )^2\right] \\
\implies &  \underset{\substack{s \sim d^{\pi_{\theta}} \\ a \sim \pi_\theta \\
s' \sim p(\cdot | s, a)}}{\E}\left[(\hat{v}_{\phi}(s') - v(s'))\nabla_\phi \hat{v}_{\phi}(s')\right] = 0 \\
\implies &  \underset{\substack{s \sim d^{\pi_{\theta}} \\ a \sim \pi_\theta \\
s' \sim p(\cdot | s, a)}}{\E}\left[(\hat{v}_{\phi}(s') - v(s'))\nabla_\theta \log \pi_\theta (a | s)\right] = 0 \\
\implies &  \underset{\substack{s \sim d^{\pi_{\theta}} \\ a \sim \pi_\theta \\
s' \sim p(\cdot | s, a)}}{\E}\left[\hat{v}_{\phi}(s')\nabla_\theta \log \pi_\theta (a | s)\right] = \underset{\substack{s \sim d^{\pi_{\theta}} \\ a \sim \pi_\theta \\
s' \sim p(\cdot | s, a)}}{\E}\left[v(s')\nabla_\theta \log \pi_\theta (a | s)\right].
\end{align*}

Combine the above results with Theorem~\ref{thm_rpg}, we have
\begin{align*}
\nabla_\theta J(\theta)
&= \underset{\substack{s \sim d^{\pi_{\theta}} \\ a \sim \pi_\theta \\
s' \sim p(\cdot|s,a)}}{\E}\left[\nabla_\theta r(s, f_\theta(\epsilon; s))|_{\epsilon = g_\theta(a; s)} + \gamma v(s')\nabla_\theta \log \pi_{\theta}(a|s)\right] \\
&= \underset{\substack{s \sim d^{\pi_{\theta}} \\ a \sim \pi_\theta \\ s' \sim p(\cdot|s,a)}}{\E}\left[\nabla_\theta \hat{r}_{\omega}(s, f_\theta(\epsilon; s))|_{\epsilon = g_\theta(a; s)} + \gamma \hat{v}_{\phi}(s')\nabla_\theta \log \pi_{\theta}(a|s)\right].
\end{align*}
\end{proof}

\subsection{Other Related Theorems}

\begin{theorem}[Reparameterization Policy Gradient Theorem]\label{thm_repara}
Suppose that the MDP satisfies Assumption~\ref{assum1} and \ref{assum2}, then
\begin{align*}
\nabla_{\theta} J(\theta)
= \int d^{\pi_{\theta}}\!(s) p(\eps) \nabla_{\theta} f_{\theta}(\eps;s) \nabla_{a} q_{\pi_{\theta}}(s, a)|_{a=f_{\theta}(\eps;s)} \de \eps \de s.
\end{align*}
\end{theorem}

\begin{proof}

By the policy gradient theorem, we have
\begin{align*}
\nabla_{\theta} J(\theta)
= \int d^{\pi_{\theta}}(s) \pi_{\theta}(a|s) q_{\pi_{\theta}}(s,a) \nabla_{\theta} \log{\pi_{\theta}(a|s)} \de a \de s.
\end{align*}

Thus
\begin{align*}
&\nabla_{\theta} J(\theta) \\
=& \int d^{\pi_{\theta}}(s) \pi_{\theta}(a|s) q_{\pi_{\theta}}(s,a) \nabla_{\theta} \log{\pi_{\theta}(a|s)} \de a \de s \\
=& \int d^{\pi_{\theta}}(s) \left( \int q_{\pi_{\theta}}(s,a) \nabla_{\theta} \pi_{\theta}(a|s) \de a \right) \de s \\
=& \int d^{\pi_{\theta}}(s) \left[ \int \nabla_{\theta} ( q_{\pi_{\theta}}(s,a) \pi_{\theta}(a|s) ) \de a - \int \pi_{\theta}(a|s) \nabla_{\theta} q_{\pi_{\theta}}(s,a) \de a \right] \de s \\
=& \int d^{\pi_{\theta}}(s) \left[\nabla_{\theta} \left(\int q_{\pi_{\theta}}(s,a) \pi_{\theta}(a|s) \de a \right) - \int \pi_{\theta}(a|s) \nabla_{\theta} q_{\pi_{\theta}}(s,a) \de a \right] \de s \\
=& \int d^{\pi_{\theta}}(s) \left[\nabla_{\theta} \left(\int p(\eps) q_{\pi_{\theta}}(s,f_{\theta}(\eps; s)) \de \eps \right) - \int p(\eps) \nabla_{\theta} q_{\pi_{\theta}}(s,a)|_{a=f_{\theta}(\eps; s)} \de \eps \right] \de s \text{\quad (by reparameterization)} \\
=& \int d^{\pi_{\theta}}(s) \left[\int p(\eps) \left(\nabla_{\theta} f_{\theta}(\eps; s) \nabla_{a} q_{\pi_{\theta}}(s,a)|_{a=f_{\theta}(\eps; s)} + \nabla_{\theta} q_{\pi_{\theta}}(s,a)|_{a=f_{\theta}(\eps; s)} \right) \de \eps - \int p(\eps) \nabla_{\theta} q_{\pi_{\theta}}(s,a)|_{a=f_{\theta}(\eps; s)} \de \eps \right] \de s \\
=& \int d^{\pi_{\theta}}(s) \; p(\eps) \nabla_{\theta} f_{\theta}(\eps; s) \nabla_{a} q_{\pi_{\theta}}(s,a)|_{a=f_{\theta}(\eps; s)} \de \eps \de s.
\end{align*}
\end{proof}

\begin{remark}
This theorem is a direct application of reparameterization to the gradient of the policy objective. It is understood to be known but is not formally presented or derived for policy gradients in any existing work. 
We include it here for completeness as well as a useful theoretical tool.
\end{remark}

\begin{corollary}[Deterministic Policy Gradient Theorem]\label{cor_dpg}
Suppose that the MDP satisfies Assumption~\ref{assum1} and \ref{assum2}, for a deterministic policy $\mu_{\theta}(s)$, we have
\begin{align*}
\nabla_{\theta} J(\theta) 
= \int d^{\mu_{\theta}}(s) \nabla_{\theta} \mu_{\theta}(s) \nabla_{a} q_{\mu_{\theta}}(s,a)|_{a=\mu_{\theta}(s)} \de s.
\end{align*}
\end{corollary}

\begin{proof}
Let $p(\eps)$ be the delta function. Thus $\int_{\eps} p(\eps) f_{\theta}(\eps; s) \de \eps = f_{\theta}(0;s)$. Furthermore, let $f_{\theta}(0;s)=\mu_{\theta}(s)$. By Theorem~\ref{thm_repara},
\begin{align*}
\nabla_{\theta} J(\theta) 
=& \int d^{\pi_{\theta}}(s) \left(\int p(\eps) \nabla_{\theta} f_{\theta}(\eps; s) \nabla_{a} q_{\pi_{\theta}}(s,a)|_{a=f_{\theta}(\eps; s)} \de \eps \right) \de s \\
=& \int d^{\pi_{\theta}}(s) \nabla_{\theta} f_{\theta}(0;s) \nabla_{a} q_{\pi_{\theta}}(s,a)|_{a=f_{\theta}(0;s)} \de s \\
=& \int d^{\pi_{\theta}}(s) \nabla_{\theta} \mu_{\theta}(s) \nabla_{a} q_{\pi_{\theta}}(s,a)|_{a=\mu_{\theta}(s)} \de s.
\end{align*}
\end{proof}

\begin{remark}
Both DPG and DDPG are proposed based on this theorem, which is first proved by~\citet{silver2014deterministic}. It is limited to deterministic policies and thus can be deduced as a corollary of Theorem~\ref{thm_repara}, which is applicable to stochastic policies as well.
\end{remark}

\begin{corollary}[Entropy-regularized Reparameterization Policy Gradient Theorem]\label{cor_sac}
Consider the entropy-regularized values $v_{\pi_{\theta}}(s_0) = \E_{\pi}\left[\sum_{t=0}^{\infty} \gamma^{t}(r(s_t, a_t) + \alpha \gH(\pi_{\theta}(\cdot | s_t)))\right]$ and $q_{\pi_{\theta}}(s_0, a_0) = \E_{\pi} \left[\sum_{t=0}^{\infty} \gamma^{t} r(s_t, a_t) + \alpha \sum_{t=1}^{\infty} \gamma^{t} \gH(\pi_{\theta}(\cdot | s_t)) \right]$, where $\gH(p) = - \int_{x} p(x) \log{p(x)} \de x$ is the differential entropy for probability density function $p(x)$, and $\alpha$ is a positive constant. Suppose that the MDP satisfies Assumption~\ref{assum1} and \ref{assum2}, then
\begin{align*}
\nabla_{\theta} J(\theta)
= \int d^{\pi_{\theta}}(s) p(\eps) \left[\nabla_{\theta} f_{\theta}(\eps; s) \nabla_{a} q_{\pi_{\theta}}(s, a)|_{a=f_{\theta}(\eps; s)} - \alpha \nabla_{\theta} \log{\pi_{\theta}(f_{\theta}(\eps; s)|s)} \right] \de \eps \de s.
\end{align*}
\end{corollary}

\begin{proof}
By definition, we have
\begin{align*}
v_{\pi_{\theta}}(s) &= \E_{\pi}\left[\sum_{t=0}^{\infty} \gamma^{t}(r(s_t, a_t) + \alpha \gH(\pi_{\theta}(\cdot | s_t))) \mid s_0=s \right], \\
q_{\pi_{\theta}}(s,a) &= \E_{\pi}\left[\sum_{t=0}^{\infty} \gamma^{t} r(s_t, a_t) + \alpha \sum_{t=1}^{\infty} \gamma^{t} \gH(\pi_{\theta}(\cdot | s_t)) \mid s_0=s, a_0=a \right],
\end{align*}
where $\gH(p) = - \int_{x} p(x) \log{p(x)} \de x$ is the differential entropy for probability density function $p(x)$.

Then $v_{\pi_{\theta}}$ and $q_{\pi_{\theta}}$ are connected by
\begin{align*}
v_{\pi_{\theta}}(s)
&= \E_{\pi}[q_{\pi_{\theta}}(s,a)] + \alpha \gH(\pi_{\theta}(\cdot | s)) \\
&= \int \pi_{\theta}(a|s) (q_{\pi_{\theta}}(s,a) - \alpha \log \pi_{\theta}(a|s)) \de a \\
&= \int p(\eps) \left(q_{\pi_{\theta}}(s,f_{\theta}(\eps; s)) - \alpha \log{\pi_{\theta}(f_{\theta}(\eps; s) | s)} \right) \de \eps,
\end{align*}
and the Bellman equation for $q_{\pi_{\theta}}$ is
\begin{align*}
q_{\pi_{\theta}}(s,a) 
&= \E_{\pi}\left[r(s, a) + \gamma(q_{\pi_{\theta}}(s',a') + \alpha \gH(\pi_{\theta}(\cdot | s'))) \right] \\
&= r(s,a) + \gamma \E[v_{\pi_{\theta}}(s')] 
= r(s,a) + \gamma \int p(s' | s,a) v_{\pi_{\theta}}(s') \de s'.
\end{align*}

Then
\begin{align*}
& \nabla_{\theta} v_{\pi_{\theta}}(s) \\
=& \nabla_{\theta} \left( \int \pi_{\theta}(a|s) q_{\pi_{\theta}}(s,a) \de a \right) + \alpha \nabla_{\theta} \gH(\pi_{\theta}(\cdot | s)) \\
=& \nabla_{\theta} \left( \int p(\eps) q_{\pi_{\theta}}(s,f_{\theta}(\eps; s)) \de \eps \right) + \alpha \nabla_{\theta} \gH(\pi_{\theta}(\cdot | s)) \\
=& \int p(\eps) \nabla_{\theta} q_{\pi_{\theta}}(s,f_{\theta}(\eps; s)) \de \eps + \alpha \nabla_{\theta} \gH(\pi_{\theta}(\cdot | s)) \\
=& \int p(\eps) \left(\nabla_{\theta} f_{\theta}(\eps; s) \nabla_{a} q_{\pi_{\theta}}(s,a)|_{a=f_{\theta}(\eps; s)} + \nabla_{\theta} q_{\pi_{\theta}}(s,a)|_{a=f_{\theta}(\eps; s)} \right) \de \eps + \alpha \nabla_{\theta} \gH(\pi_{\theta}(\cdot | s)) \\
=& \int p(\eps) \left(\nabla_{\theta} f_{\theta}(\eps; s) \nabla_{a} q_{\pi_{\theta}}(s,a)|_{a=f_{\theta}(\eps; s)} + \gamma \int p(s' | s,f_{\theta}(\eps; s)) \nabla_{\theta} v_{\pi_{\theta}}(s') \de s' \right) \de \eps + \alpha \nabla_{\theta} \gH(\pi_{\theta}(\cdot | s)) \\
=& \int p(\eps) \nabla_{\theta} f_{\theta}(\eps; s) \nabla_{a} q_{\pi_{\theta}}(s,a)|_{a=f_{\theta}(\eps; s)} \de \eps + \gamma \int p(\eps) p(s' | s,f_{\theta}(\eps; s)) \nabla_{\theta} v_{\pi_{\theta}}(s') \de \eps \de s' + \alpha \nabla_{\theta} \gH(\pi_{\theta}(\cdot | s)) \\
=& \int p(\eps) \nabla_{\theta} f_{\theta}(\eps; s) \nabla_{a} q_{\pi_{\theta}}(s,a)|_{a=f_{\theta}(\eps; s)} \de \eps + \gamma \int p(s \to s', 1, \pi_{\theta}) \nabla_{\theta} v_{\pi_{\theta}}(s') \de s' + \alpha \nabla_{\theta} \gH(\pi_{\theta}(\cdot | s)),
\end{align*}
where $p(s \to s', 1, \pi_{\theta}) = \int p(\eps) p(s' | s,f_{\theta}(\eps; s)) \de \eps$.

Now iterating this formula we have
\begin{align*}
& \nabla_{\theta} v_{\pi_{\theta}}(s) \\
=& \alpha \nabla_{\theta} \gH(\pi_{\theta}(\cdot | s)) + \int p(\eps) \nabla_{\theta} f_{\theta}(\eps; s) \nabla_{a} q_{\pi_{\theta}}(s,a)|_{a=f_{\theta}(\eps; s)} \de \eps + \gamma \int p(s \to s', 1, \pi_{\theta}) \nabla_{\theta} v_{\pi_{\theta}}(s') \de s' \\
=& \alpha \nabla_{\theta} \gH(\pi_{\theta}(\cdot | s)) + \int p(\eps) \nabla_{\theta} f_{\theta}(\eps; s) \nabla_{a} q_{\pi_{\theta}}(s,a)|_{a=f_{\theta}(\eps; s)} \de \eps + \gamma \int p(s \to s', 1, \pi_{\theta}) \\
& \left(\alpha \nabla_{\theta} \gH(\pi_{\theta}(\cdot | s')) + \int p(\eps') \nabla_{\theta} f_{\theta}(\eps'; s') \nabla_{a} q_{\pi_{\theta}}(s',a)|_{a=f_{\theta}(\eps'; s')} \de \eps' + \gamma \int p(s' \to s'', 1, \pi_{\theta}) \nabla_{\theta} v_{\pi_{\theta}}(s'') \de s''\right) \de s' \\
=& \alpha \nabla_{\theta} \gH(\pi_{\theta}(\cdot | s)) + \int p(\eps) \nabla_{\theta} f_{\theta}(\eps; s) \nabla_{a} q_{\pi_{\theta}}(s,a)|_{a=f_{\theta}(\eps; s)} \de \eps \\
&+ \alpha \gamma \int p(s \to s', 1, \pi_{\theta}) \nabla_{\theta} \gH(\pi_{\theta}(\cdot | s')) \de s' + \gamma \int p(s \to s', 1, \pi_{\theta}) p(\eps') \nabla_{\theta} f_{\theta}(\eps'; s') \nabla_{a} q_{\pi_{\theta}}(s',a)|_{a=f_{\theta}(\eps'; s')} \de \eps' \de s' \\
&+ \gamma^2 \int p(s \to s', 1, \pi_{\theta}) p(s' \to s'', 1, \pi_{\theta}) \nabla_{\theta} v_{\pi_{\theta}}(s'') \de s'' \de s' \\
=& \alpha \nabla_{\theta} \gH(\pi_{\theta}(\cdot | s)) + \int p(\eps) \nabla_{\theta} f_{\theta}(\eps; s) \nabla_{a} q_{\pi_{\theta}}(s,a)|_{a=f_{\theta}(\eps; s)} \de \eps \\
&+ \alpha \gamma \int p(s \to s', 1, \pi_{\theta}) \nabla_{\theta} \gH(\pi_{\theta}(\cdot | s')) \de s' + \gamma \int p(s \to s', 1, \pi_{\theta}) p(\eps') \nabla_{\theta} f_{\theta}(\eps'; s') \nabla_{a} q_{\pi_{\theta}}(s',a)|_{a=f_{\theta}(\eps'; s')} \de \eps' \de s' \\
&+ \gamma^2 \int p(s \to s', 2, \pi_{\theta}) \nabla_{\theta} v_{\pi_{\theta}}(s') \de s' \\
=& \hdots\\
=& \int \sum_{t=0}^{\infty} \gamma^{t} p(s \to s', t, \pi_{\theta}) \left(\alpha \nabla_{\theta} \gH(\pi_{\theta}(\cdot | s')) + \int p(\eps') \nabla_{\theta} f_{\theta}(\eps'; s') \nabla_{a} q_{\pi_{\theta}}(s',a)|_{a=f_{\theta}(\eps'; s')} \de \eps' \right) \de s',
\end{align*}
where $p(s \to s'', t+1, \pi_{\theta}) = \int p(s \to s', t, \pi_{\theta}) p(s' \to s'', 1, \pi_{\theta}) \de s'$.

Furthermore,
\begin{align*}
\nabla_{\theta} \gH(\pi_{\theta}(\cdot | s))
= - \nabla_{\theta} \int \pi_{\theta}(a|s) \log{\pi_{\theta}(a|s)} \de a 
= - \nabla_{\theta} \int p(\eps) \log{\pi_{\theta}(f_{\theta}(\eps; s) | s)} \de \eps
= - \int p(\eps) \nabla_{\theta} \log{\pi_{\theta}(f_{\theta}(\eps; s) | s)} \de \eps.
\end{align*}

Then
\begin{align*}
\nabla_{\theta} v_{\pi_{\theta}}(s)
=& \int \sum_{t=0}^{\infty} \gamma^{t} p(s \to s', t, \pi_{\theta}) \left(\alpha \nabla_{\theta} \gH(\pi_{\theta}(\cdot | s')) + \int p(\eps') \nabla_{\theta} f_{\theta}(\eps'; s') \nabla_{a} q_{\pi_{\theta}}(s',a)|_{a=f_{\theta}(\eps'; s')} \de \eps' \right) \de s' \\
=& \int \sum_{t=0}^{\infty} \gamma^{t} p(s \to s', t, \pi_{\theta}) p(\eps') \left(\nabla_{\theta} f_{\theta}(\eps'; s') \nabla_{a} q_{\pi_{\theta}}(s',a)|_{a=f_{\theta}(\eps'; s')} - \alpha \nabla_{\theta} \log{\pi_{\theta}(f_{\theta}(s,\eps') | s)} \right) \de \eps' \de s'.
\end{align*}

Finally,
\begin{align*}
\nabla_{\theta} J(\theta)
=& \nabla_{\theta} \int p_0(s) v_{\pi_{\theta}}(s) \de s = \int p_0(s) \nabla_{\theta} v_{\pi_{\theta}}(s) \de s \\
=& \int \sum_{t=0}^{\infty} \gamma^{t}  p_0(s) p(s \to s', t, \pi_{\theta}) p(\eps) \left(\nabla_{\theta} f_{\theta}(\eps; s') \nabla_{a} q_{\pi_{\theta}}(s',a)|_{a=f_{\theta}(\eps; s')} - \alpha \nabla_{\theta} \log{\pi_{\theta}(f_{\theta}(\eps; s) | s)} \right) \de \eps \de s' \de s \\
=& \int d^{\pi_{\theta}}(s) \; p(\eps) \left(\nabla_{\theta} f_{\theta}(\eps; s) \nabla_{a} q_{\pi_{\theta}}(s,a)|_{a=f_{\theta}(\eps; s)} - \alpha \nabla_{\theta} \log{\pi_{\theta}(f_{\theta}(\eps; s) | s)} \right)  \de \eps \de s.
\end{align*}
\end{proof}

\begin{remark}
There is a similar result presented by~\cite{haarnoja2018soft} for SAC. They obtain it by minimizing the Kullback-Leibler divergence between the new policy and the policy derived from the exponential of the soft Q-function. However, we derive it by directly minimizing the objective with the help of the RP gradient. This corollary provides an alternative way to understand SAC. 
\end{remark}

\section{Hyper-parameter Settings for Simulation Tasks}

All our experiments were performed only with CPUs (AMD EPYC 7601 32-Core Processor).

\subsection{LQG}

The parameters used for the LQG task are
\begin{align*}
A=
\begin{bmatrix}
0.01 & 0\\
0 & 0.01
\end{bmatrix}
;\quad
B=
\begin{bmatrix}
1e-4 & 0\\
0 & 1e-4
\end{bmatrix}
;\quad
Q=
\begin{bmatrix}
1 & 0\\
0 & 1
\end{bmatrix}
;\quad
R=
\begin{bmatrix}
1 & 0\\
0 & 1
\end{bmatrix}
;\quad
\Sigma=
\begin{bmatrix}
0.1 & 0\\
0 & 0.1
\end{bmatrix}
;\quad
\textbf{s}_0 = [0.5, 0.5].
\end{align*}

The discount factor is $\gamma=0.99$. The number of steps for each episode is 100. The policy parameter is chosen randomly to be $\theta=[-1.1104430687690852, -1.3649958298432607]$.

\subsection{MuJoCo} 

\begin{table}[thbp]
\caption{The hyper-parameter settings for PPO and RPG on MuJoCo tasks.} \label{tb_hyper_ppo_rpg}
\begin{center}
\begin{tabular}{lcc}
\toprule
\bf{Hyper-parameter} & \bf{PPO} & \bf{RPG} \\
\midrule
Policy network LR & $3 \times 10^{-4}$ & $3 \times 10^{-4}$ \\
Value network LR & $10^{-3}$ & $10^{-3}$ \\
Reward network LR & None & $10^{-3}$ \\
Hidden layers & [64, 64] & [64, 64] \\
Optimizer & Adam & Adam \\
Time-steps per iteration & 2048 & 2028 \\
Number of epochs & 10 & 10 \\
Mini-batch size & 64 & 64 \\ 
Discount factor ($\gamma$) & 0.99 & 0.99 \\
GAE parameter ($\lambda$) & 0.95 & 0.95 \\
PPO Clipping ($\eps$) & 0.2 & 0.2 \\
Target KL divergence & 0.01 & 0.01 \\
State Clipping & [-10, 10] & [-10, 10] \\
Gradient clipping & 2 & 2 \\
\bottomrule
\end{tabular}
\end{center}
\end{table}

\begin{algorithm}[b]
\caption{PPO} \label{algo_ppo}
\begin{algorithmic}
    \STATE Input: initial policy parameters $\theta$, initial value estimate parameters $\phi$.
    \FOR{$k=0,1,2,\dots$}
        \STATE Collect set of trajectories $\mathcal{D}=\{\tau_{i}\}$ by running policy $\pi_{\theta}$ in the environment. \\
        \STATE Compute returns $G_t$. \\
        \STATE Compute advantage estimates $H_t = H_t^{\text{GAE}(\lambda)}$.
        \STATE Normalize $H_{t}$ using the sample mean and standard deviation of all advantage estimates.
        \FOR{epoch $=0,1,2,\dots$}
        \STATE Shuffle and slice trajectories $\mathcal{D}$ into mini-batches.
            \FOR{each mini-batch $B$}
                \STATE Set $\hat{\rho}_t(\theta)$ according to Equation~\ref{eq_clip_new}.
                \STATE Update policy parameters $\theta$ by maximizing the objective: $\E_{B}[\hat{\rho}_t(\theta) H_{t}]$.
                \STATE Update value estimate parameters $\phi$ by minimizing: $\E_{B}[(\hat{v}_{\phi}(S_{t})-G_t)^{2}]$.
            \ENDFOR
        \ENDFOR
    \ENDFOR
\end{algorithmic}
\end{algorithm}

\section{Real-Robot Task Description}

For the real-robot experiment in this paper, we adopted the \textit{UR-Reacher-2} task from Mahmood et al.\ (2018).
The task runs on a UR5 robot arm from Universal Robots.
By moving only two joints of the robot---the base and the elbow joints---an agent can move the robot around in two dimensions above a table to which the base joint is attached.
The elbow joint connects the base joint to the fingertip.

Each episode lasts for 100 time steps of 40ms duration or 4 seconds in total.
At the start of each episode the fingertip is reset to a fixed position above the robot, and a random target is drawn uniformly from a rectangular area around the fingertip.
The goal of the agent is to get the fingertip as close and as fast as possible to the target by moving the base and elbow joints.
The reward at each time step has two terms.
The first term is the negative of the distance from fingertip to target.
The second term rewards the agent equal to a Gaussian function that has its maximum value when the fingertip is on the target and reduces gradually as the fingertip moves away.
In other words, the Gaussian function is centered at the target and evaluated at the distance from fingertip to target.
This second reward term is used to encourage precision when fingertip and target are close by.
The observation vector includes the angular position and velocity of the base and elbow joints, the vector from the fingertip to the target, and the previous action.
The action vector contains the target angular velocity that the base and elbow joints should be set to.

In the original environment, whenever the learning agent pushed the fingertip outside a 2-dimensional rectangular boundary above the robot, a manually written script would move the fingertip to be inside the boundary.
To reduce the frequency of scripted position corrections, we relaxed their movement constraints to a much larger area similarly to Farrahi and Mahmood (2020).
The fingertip can now move freely as long as it is not too close to the table.
The base and elbow joints can now rotate farther as well.
We additionally made a few changes to the environment's code, without changing the task setup, to eliminate the protective stops that occurred when operating the joints at high speeds.